\documentclass[final,5p,times,twocolumn]{elsarticle}

\newcommand{\textnormalit}[1]{\textnormal{\it #1}}

\usepackage{graphics}
\usepackage{graphicx}
\usepackage{amsfonts}
\usepackage{amssymb,amsmath,subfigure}
\usepackage{multirow}
\usepackage{algorithmic}
\usepackage{lscape}
\usepackage{longtable}
\usepackage{array}
\usepackage{longtable}
\usepackage{multicol}
\usepackage{afterpage}
\usepackage{moresize}
\usepackage[none]{hyphenat}
\usepackage{algorithm2e}
\usepackage{url}

\newcolumntype{H}{>{\setbox0=\hbox\bgroup}c<{\egroup}@{}}

\journal{Applied Soft Computing}

\begin{document}

\begin{frontmatter}
\title{Learning Fuzzy Controllers in Mobile Robotics with Embedded Preprocessing}
\author{I.~Rodr\'iguez-Fdez\corref{irf}}
\ead{ismael.rodriguez@usc.es}
\author{M.~Mucientes\corref{ryc}}
\ead{manuel.mucientes@usc.es}
\author{A.~Bugar\'in\corref{}}
\ead{alberto.bugarin.diz@usc.es}

\cortext[irf]{Corresponding author. Tel.: +34 881816392.}

\address{Centro de Investigaci\'on en Tecnolox\'ias da Informaci\'on (CITIUS), Universidade de Santiago de Compostela, SPAIN}

\begin{abstract}
The automatic design of controllers for mobile robots usually requires two stages. In the first stage, sensorial data are preprocessed or transformed into high level and meaningful values of variables which are usually defined from expert knowledge. In the second stage, a machine learning technique is applied to obtain a controller that maps these high level variables to the control commands that are actually sent to the robot. This paper describes an algorithm that is able to embed the preprocessing stage into the learning stage in order to get controllers directly starting from sensorial raw data with no expert knowledge involved. Due to the high dimensionality of the sensorial data, this approach uses Quantified Fuzzy Rules (QFRs), that are able to transform low-level input variables into high-level input variables, reducing the dimensionality through summarization. The proposed learning algorithm, called Iterative Quantified Fuzzy Rule Learning (IQFRL), is based on genetic programming. IQFRL is able to learn rules with different structures, and can manage linguistic variables with multiple granularities. The algorithm has been tested with the implementation of the wall-following behavior both in several realistic simulated environments with different complexity and on a \textit{Pioneer 3-AT} robot in two real environments. Results have been compared with several well-known learning algorithms combined with different data preprocessing techniques, showing that IQFRL exhibits a better and statistically significant performance. Moreover, three real world applications for which IQFRL plays a central role are also presented: path and object tracking with static and moving obstacles avoidance.
\end{abstract}

\begin{keyword}
mobile robotics \sep Quantified Fuzzy Rules \sep Iterative Rule Learning \sep Genetic Fuzzy System
\end{keyword}

\end{frontmatter}

\section{Introduction}
The control architecture of mobile robots usually includes a number of behaviors that are implemented as controllers, which are able to solve specific tasks such as motion planning, following a moving object, wall-following, avoiding collisions, etc. in real time. These behaviors are implemented as controllers whose outputs at each time point (control commands) depend on both the internal state of the robot and the environment in which it evolves. The robot sensors (e.g. laser range finders, sonars, cameras, etc.) are used in order to obtain the augmented state of the robot (internal state and environment). When the robot operates in real environments, both the data obtained by these sensors and the internal state of the robot present uncertainty or noise. Therefore, the use of mechanisms that manage them properly is necessary. The use of fuzzy rules is convenient to cope with this uncertainty, since it combines the interpretability and expressiveness of the rules with the ability of fuzzy logic for 
representing uncertainty.

The first step for designing controllers for mobile robots consists of the preprocessing of the raw sensor data: the low-level input variables obtained by the sensors are transformed into high-level variables that are significant for the behavior to be learned. Usually, expert knowledge is used for the definition of these high-level variables and the mapping from the sensorial data. After this preprocessing stage, machine learning algorithms can be used to automatically obtain the mapping from the high-level input variables to the robot control commands. This paper describes an algorithm that is able to perform the preprocessing stage embedded in the learning stage, thus avoiding the use of expert knowledge. Therefore, the mapping between low-level and high-level input variables is done automatically during the learning phase of the controller.

The data provided by the sensors is of high dimensionality. For example, a robot equipped with two laser range finders can generate over 720 low-level variables. However, in mobile robotics it is more common to work with sets or groupings of these variables, (e.g. ``frontal sector'') that are much more significant and relevant for the behavior. As a result, it is necessary to use a model that is capable of grouping low-level variables, thus reducing the dimensionality of the problem and providing meaningful descriptions. The model should provide propositions that are able to summarize the data with expressions like ``part of the distances in the frontal sector are high''. This kind of expressions can model the underlying knowledge in a better way than just using average, maximum or minimum values of sets of low level variables. Moreover, these expressions also include the definition of the set of low-level variables to be used. Since these propositions involve fuzzy quantifiers (e.g. ``part''), they are 
called Quantified Fuzzy Propositions (QFPs) \cite{mucientes2010_pr}. QFP provide a formal model that is capable of modeling the knowledge involved in this grouping task.

Evolutionary algorithms have some characteristics that make them suitable for learning fuzzy rules. The well-known combination of evolutionary algorithms and fuzzy logic (genetic fuzzy systems) is one of the approaches that aims to manage the balance between accuracy and interpretability of the rules \cite{cordon2004ten,herrera2008_ei}. As it was pointed out before, fuzzy rules can be composed of both conventional and QFPs (therefore, they will be referred to as QFRs). Furthermore, the transformation from low-level to high-level variables using QFPs produces a variable number of propositions in the antecedent of the rules. Therefore, genetic programming, where the structure of individuals is a tree of variable size derived from a context-free grammar, is here the most appropriate choice.

This paper describes an algorithm that is able to learn QFRs of variable structure for the design of controllers with embedded preprocessing in mobile robotics. This proposal, called Iterative Quantified Fuzzy Rule Learning (IQFRL), is based on the Iterative Rule Learning (IRL) approach and uses linguistic labels defined with unconstrained multiple granularity, i.e. without limiting the granularity levels. This proposal has been designed to solve control (regression) problems in mobile robotics having as input variables the internal state of the robot and the sensors data. Expert knowledge is only used to generate the training data for each of the situations of the task to be learned and, also, to define the context-free grammar that specifies the structure of the rules.

The main contributions of the paper are: (i) the proposed algorithm is able to learn using the state of the robot and the sensors data, with no preprocessing. Instead, the mapping between low-level variables and high-level variables is done embedded in the algorithm; (ii) the algorithm uses QFPs, a model able to summarize the low-level input data; (iii) moreover, IQFRL uses  linguistic labels with unconstrained multiple granularity. With this approach, the interpretability of the membership functions used in the resulting rules is unaffected while the flexibility of representation remains. The proposal was validated in several simulated and real environments with the wall-following behavior. Results show a better and statistically significant performance of IQFRL over several combinations of well-known learning algorithms and preprocessing techniques. The approach was also tested in three real world behaviors that were built as a combination of controllers: path tracking with obstacles avoidance, object tracking with fixed obstacles avoidance, and object tracking with moving obstacle avoidance.

The paper is structured as follows: Section \ref{Sec:related} summarizes recent work related with this proposal and Section \ref{Sec:QFR} presents the QFRs model and its advantages in mobile robotics. Section \ref{Sec:evoAlg} describes the IQFRL algorithm that has been used to learn the QFRs. Section \ref{Sec:results} presents the obtained results, and Section \ref{Sec:realWorld} shows three real world applications of IQFRL in robotics. Finally, Section \ref{Sec:conclusions} points out the most relevant conclusions.

\section{Related Work\label{Sec:related}}
The learning of controllers for autonomous robots has been dealt with by using different machine learning techniques. Among the most popular approaches can be found evolutionary algorithms \cite{bonte2010_aiai, martinez2014genetic}, neural networks \cite{umar2011_rcim} and reinforcement learning \cite{agostini2010_INSTICC, lo2014intelligent}. Also hibridations of them, like evolutionary neural networks \cite{kondo2007_asc}, reinforcement learning with evolutionary algorithms \cite{samsudin2011_asc,mabu2010_smc}, the widely used genetic fuzzy systems \cite{senthilkumar2009hybrid,mucientes2007_asc,mucientes2009_ijis,Mucientes10_eswa,kuo2009_his,khanian2009_cira,Mucientes06_sc}, or even more uncommon combinations like ant colony optimization with reinforcement learning  \cite{juang2009_ie} or differential evolution \cite{hsu2013evolutionary} or evolutionary group based particle swarm optimization \cite{juang2011_fs} have been successfully applied. Furthermore, over the last few years, mobile robotic controllers have been getting some attention as a test case for the automatic design of type-2 fuzzy logic controllers \cite{lo2014intelligent, martinez2014genetic, hsu2013evolutionary}.

An extensive use of expert knowledge is made in all of these approaches. In \cite{senthilkumar2009hybrid} 360 laser sensor beams are used as input data, and are heuristically combined into 8 sectors as inputs to the learning algorithm. On the other hand, in \cite{kondo2007_asc,mucientes2007_asc,mucientes2009_ijis,Mucientes10_eswa,kuo2009_his,Mucientes06_sc,juang2009_ie,juang2011_fs} the input variables of the learning algorithm are defined by an expert. Moreover, in \cite{mucientes2007_asc,mucientes2009_ijis,kuo2009_his,Mucientes06_sc, hsu2013evolutionary} the evaluation function of the evolutionary algorithm must be defined by an expert for each particular behavior. As in the latter case, the reinforcement learning approaches need the definition of an appropriate reward function using expert knowledge.

The approaches based on genetic fuzzy systems use different alternatives in the definition of the membership functions. In  \cite{samsudin2011_asc,senthilkumar2009hybrid,kuo2009_his} the membership functions are defined heuristically. In  \cite{mucientes2009_ijis,Mucientes10_eswa} labels have been uniformly distributed, but the granularity of each input variable is defined using expert knowledge. On the other hand, in \cite{mucientes2007_asc,khanian2009_cira,Mucientes06_sc,juang2009_ie,juang2011_fs} an approximative approach is used, i.e., different membership functions are learned for each rule, reducing the interpretability of the learned controller.

The main problem of learning behaviors using raw sensor input data is the curse of dimensionality. In \cite{agostini2010_INSTICC}, this issue has been managed from the reinforcement learning perspective, by using a probability density estimation of the joint space of states. Among all the approaches based on evolutionary algorithms, only in \cite{bonte2010_aiai} no expert knowledge has been taken into account. In this work, the number of sensors and their position are learned from a reduced number of sensors.

In \cite{mucientes2009_ifsa} a Genetic Cooperative-Competitive Learning (GCCL) approach was presented. The proposal learns knowledge bases without preprocessing raw data, but the rules involved approximative labels while the IQFRL proposal uses unconstrained multiple granularity. Moreover, in this approach it is difficult to adjust the balance between cooperation and competition, which is typical when learning rules in GCCL. As a result, the obtained rules where quite specific and the performance of the behavior was not comparable to other proposals based on expert knowledge.

\section{Quantified Fuzzy Rules (QFRs)\label{Sec:QFR}}

\subsection{QFRs for robotics}
Machine learning techniques in mobile robotics are used to obtain the mapping from inputs to outputs (control commands). In general, two categories can be established for the input variables: 

\begin{itemize}
\item High-level input variables: variables that provide, by themselves, information that is relevant and meaningful to the expert for modeling the system (e.g. the linear velocity of the robot, or the right-hand distance from the robot to a wall).

\item Low-level input variables: variables that do not provide by themselves information for the expert to model the system (e.g. a single distance measure provided by a sensor). Relevance of these variables emerge when they are grouped into more significant sets of variables. For example, the control actions cannot be decided by simply analyzing the individual distance values provided by each beam of a laser range finder, since noisy measurements or gaps between objects (very frequent in cluttered environments) may occur. Instead, more significant variables and models involving complex groupings and structures are used.
\end{itemize}

Usually, high-level variables, or sectors, consisting of a set of laser beam measures instead of the beam measures themselves (e.g., right distance, frontal distance, etc.) are used in mobile robotics. The low-level input variables are transformed into high-level input variables in a preprocessing stage previous to the learning of the controller. Traditionally, this transformation and the resulting high-level input variables are defined using expert knowledge. Doing this preprocessing automatically during the learning phase demands a model that groups the low-level input variables in an expressive and meaningful way. Within this context Quantified Fuzzy Propositions (QFPs) such as \emph{``part of the distances of the frontal sector are low''} are useful for representing relevant knowledge for the experts and therefore for performing intelligent control. Modeling with QFPs as in the previous example demands the definition of several elements: 

\begin{itemize}
\item \emph{part}: how many distances of the frontal sector must be low? 
\item \emph{frontal sector}: which beams belong to the frontal sector?
\item \emph{low}: what is the actual semantics of low?
\end{itemize}

This example clearly sets out the need to use propositions that are different from the conventional ones. The use of QFPs in robotics eliminates the need of expert knowledge in two ways: i) the preprocessing of the low-level variables can be embedded in the learning stage; ii) the definition of the high-level variables obtained from low-level variables is done automatically, also during the learning stage. In this paper QFPs are used for representing knowledge about high-level variables that are defined as the grouping of low-level variables. Conventional fuzzy propositions are also used to represent conventional high-level variables, i.e., high-level variables not related to low-level ones (e.g. velocity).

\subsection{QFRs model\label{sec:qfr}}

An example of a QFR is shown in Fig. \ref{Fig:QFR}, involving both QFPs (\ref{Eq:distProp_1}) and conventional ones (\ref{Eq:velProp}); the outputs of the rule are also fuzzy sets. In order to determine the degree to which the output of the rule will be applied, it is necessary to reason about the propositions (using, for example, the Mamdani's reasoning scheme).

\begin{figure}[tb!]
\centering \setlength{\fboxrule}{0.2mm} \fbox{\parbox{0.45\textwidth}{
\setlength{\leftmargini}{0.5cm}
 
\begin{align}
\label{Eq:distProp_1}
& \textnormal{IF}\ part\ of\ \textnormalit{distances}\ \textnormalit{of} \: \textnormalit{FRONTAL\ SECTOR} \: \textnormalit{are} \: LOW
\
\ \textnormal{and}\\
\nonumber
& \ldots \\
\label{Eq:velProp}
& \textnormalit{velocity} \: \textnormalit{is} \: \textnormalit{HIGH}\\
\nonumber
& \textnormal{THEN} \ \textnormalit{vlin} \: \textnormalit{is} \:
\textnormalit{VERY LOW} \ \ \textnormal{and} \ \ \textnormalit{vang}
\:
\textnormalit{is} \: \textnormalit{TURN LEFT}
\end{align}

}}
\caption[]{\label{Fig:QFR}An example of QFR to model the behavior of a mobile robot.}
\end{figure}

The general expression for QFPs in this case is:

\begin{equation}
\label{Eq:distProp}
d \left( h \right) \: \textnormalit{is} \: F_d^i  \: \textnormalit{in} \: Q^i \: \textnormalit{of} \: F_{\textnormalit{b}}^i
\end{equation}
 where, for each $i$=1, ..., $g_{b}^{max}$ ($g_{b}^{max}$ being the maximum possible number of sectors of distances):
\begin{itemize}
\item $d \left( h \right)$ is the signal. In this example, it represents the distance measured by beam $h$.
\item $F_d^i$ is a linguistic value for variable $d \left( h \right)$ (e.g., \textit{``low''}).
\item $Q^i$ is a (spatial, defined in the laser beam domain) fuzzy quantifier (e.g., \textit{``part''}).
\item $F_{\textnormalit{b}}^i$ is a fuzzy set in the laser beam domain (e.g., the \textit{``frontal sector''}).
\end{itemize}

Evaluation of the Degree of Fulfillment ($DOF$) for QFP (Eq. \ref{Eq:distProp}) is carried out using Zadeh's quantification model for proportional quantifiers (such as ``most of'', ``part of'', ...) \cite{zadeh1983_cma}. This model allows to consider non-persistence, partial persistence and total persistence situations for the event ``$d \left( h \right)$ is $F_d^i$'' in the range of laser beams (spatial interval $F_{\textnormalit{b}}^i$). Therefore, for the considered example, it is possible  to make a total or partial assessment on how many distances should be low, in order to decide the corresponding  control action. This is a relevant feature of this model, since it allows to consider partial, single or total fulfillment of an event within the laser beams set.

The number of analyzed sectors of distances and their definition may vary for each of the rules. There can be very generic rules that only need to evaluate a single sector consisting of many laser beams, while other rules may need a finer granularity, with more specific laser sectors. Moreover, the rules may require a mix of QFPs and standard fuzzy propositions (for conventional high-level variables). Therefore, the automatic learning of QFRs demands an algorithm with the capability of managing rules with different structures.

\section{Iterative Quantified Fuzzy Rule Learning of Controllers\label{Sec:evoAlg}}

\subsection{Evolutionary learning of Knowledge Bases}
Evolutionary learning methods follow two approaches in order to encode rules within a population of individuals \cite{herrera2008_ei,cordón2001_ws}:
\begin{itemize}
 \item Pittsburgh approach: each individual represents the entire rule base.
 \item Michigan, IRL \cite{cordón2001_fss}, and GCCL \cite{greene1993_ml}: each individual codifies a rule. The learned rule base is the result of combining several individuals. The way in which the individuals interact during the learning process defines these three different approaches. 
\end{itemize}

The discussion is focused on those approaches for which an individual represents a rule, discarding the Michigan approach as it is used in reinforcement learning problems in which the reward from the environment needs to be maximized \cite{Eiben03_book}. Therefore, the IRL and GCCL approaches are analyzed.

In the IRL approach, the individuals compete among them but only a single rule is learned for each run (epoch) of the evolutionary algorithm. After each sequence of iterations, the best rule is selected and added to the final rule base. The selected rule must be penalized in order to induce niche formation in the search space. A common way to penalize the obtained rules is to delete the training examples that have been covered by the set of rules in the final rule base. The final step of the IRL approach is to check whether the obtained set of rules is a complete knowledge base. In the case it is not, the process is repeated. A weak point of this approach is that the cooperation among rules is not taken into account when a rule is evaluated. For example, a new rule could be added to the final rule base, deteriorating the behavior of the whole rule base over a set of examples that were already covered. The cooperation among rules can be improved with a posterior rules selection process.

In the GCCL approach the entire population codifies the rule base. That is, rules evolve together but competing among them to obtain the higher fitness. For this type of algorithm it is fundamental to include a mechanism to maintain the diversity of the population (niche induction). This mechanism must warrant that individuals of the same niche compete among themselves, but also has to avoid deleting those weak individuals that occupy a niche that remains uncovered. This is usually done using token competition \cite{cordón2001_ws}. 

Although GCCL works well for classification problems \cite{mucientes2010_pr}, the same does not occur for regression problems \cite{mucientes2009_ifsa}, mostly due to the difficulty of achieving in this realm an adequate balance between cooperation and competition. It is frequent in regression that an individual tries to capture examples seized by other individual, improving the performance on many of the examples, but decreasing the accuracy on a few ones. In subsequent iterations, new and more specific individuals replace the rule that was weakened. As a result, the individuals improve their individual fitness, but the performance of the knowledge base does not increase. In particular, for mobile robotics, the obtained knowledge bases over-fit the training data due to a \emph{polarization} effect of the rule base: few very general rules and many very specific rules. Moreover, many times, the errors of the individual rules compensate each other, generating a good output of the rule base over the training 
data, but not on test data.

This proposal, called IQFRL (Iterative Quantified Fuzzy Rule Learning), is based on IRL. The learning process is divided into epochs (set of iterations), and at the end of each epoch a new QFR (Sec. \ref{sec:qfr}) is obtained. The following sections describe each of the stages of the algorithm (Fig. \ref{Alg:IQFRL}).

\begin{figure}[tb!]
\begin{algorithmic}[1]
\STATE $\textnormalit{KB}_{\textnormalit{cur}} := \varnothing$
\REPEAT

\STATE $it := 0$
\STATE $equal_{ind} := 0$
    \STATE Initialization
    \STATE Evaluation
    \REPEAT
	\STATE Selection
	\STATE Crossover and Mutation
	\STATE Evaluation
	\STATE Replacement
	 \IF{$best^{it-1}_{ind} = best^{it}_{ind}$}
	\STATE $equal_{ind} := equal_{ind}\ +\ 1$
    \ELSE
	\STATE $equal_{ind} := 0$
    \ENDIF
    \STATE $it := it\ +\ 1$
    \UNTIL{$\left( it \geq it_{min} \wedge equal_{ind} \geq it_{check} \right) \vee \left( it \geq it_{max} \right) $}
    \STATE \label{Eq:finalkb} $\textnormalit{KB}_{\textnormalit{cur}} := \textnormalit{KB}_{\textnormalit{cur}} \cup best_{ind}$
    \STATE \label{Eq:remained} $uncov_{ex} := uncov_{ex}\ -\ cov_{ex}$
\UNTIL{$uncov_{ex} = \varnothing$}
\end{algorithmic}
\caption[]{\label{Alg:IQFRL}IQFRL algorithm.}
\end{figure}

\subsection{Examples and Grammar}

\begin{figure}[tb!]
\centering
\includegraphics[width=0.3\textwidth]{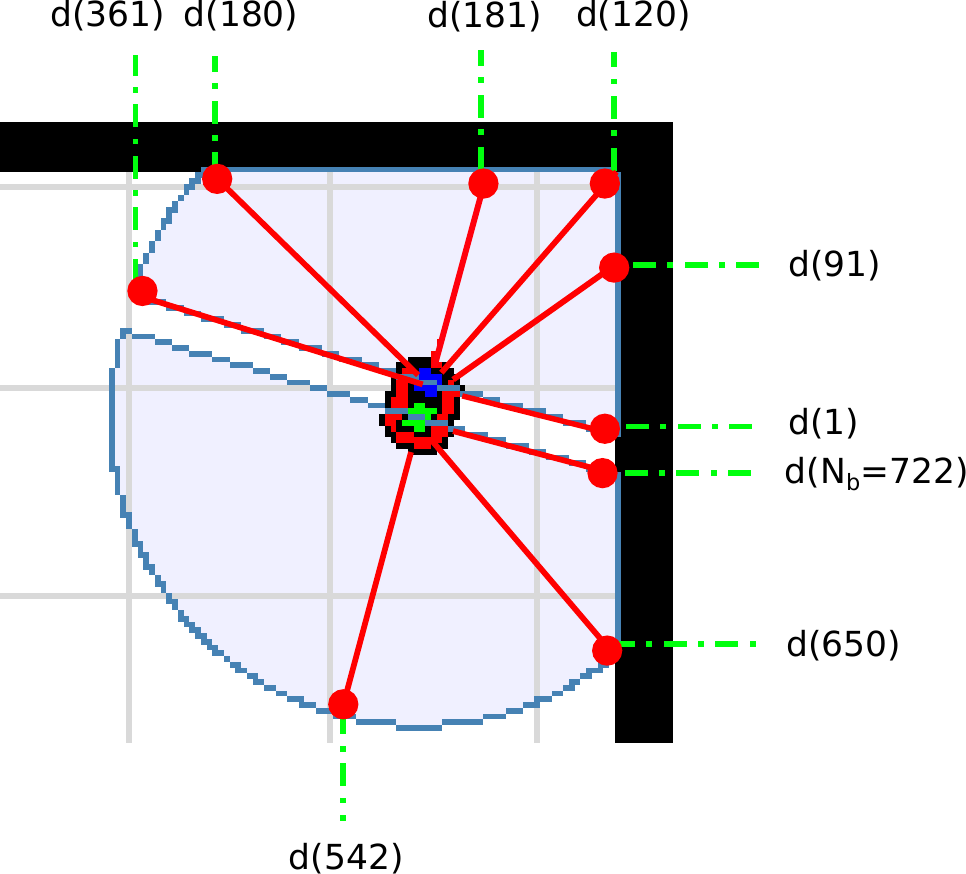}
\caption{\label{Fig:beams}Some of the distances measured by a robot equipped with two laser range finders.}
\end{figure}

The learning process is based on a set of training examples. In mobile robotics, each example can be composed of several variables that define the state of the robot (position, orientation, linear and angular velocity, etc.), and the data measured by the sensors. If the robot is equipped with laser range finders, the sensors data are vectors of distances. A laser range finder provides the distances to the closest obstacle in each direction (Fig. \ref{Fig:beams}) with a given angular resolution (number of degrees between two consecutive beams). In this paper, each example $e^l$ is represented by a tuple: 
\begin{equation}
\label{Eq:example}
e^l = \left( d\left( 1 \right), \: \ldots,  \: d\left( N_b\right), \: \textnormalit{velocity}, \: \textnormalit{vlin}, \: \textnormalit{vang} \right)
\end{equation}
where $d\left( h \right)$ is the distance measured by beam $h$, $N_b$ is the number of beams (e.g. 722 for a robot equipped with two Sick LMS200 laser range scanners as in Fig. \ref{Fig:beams}), $\textnormalit{velocity}$ is the measured linear velocity of the robot, and $\textnormalit{vlin}$ and $\textnormalit{vang}$ are the output variables (control commands for the linear and angular velocities respectively).

The individuals in the population include both conventional propositions and QFPs (Sec. \ref{sec:qfr}). Also, the number of relevant inputs can be different. Therefore, genetic programming is the most appropriate approach, as each individual is a tree of variable size.  In order to generate valid individuals of the population, and to produce right structures for the individuals after crossover and mutation, some constraints have to be added. With a context-free grammar all the valid structures of a tree (genotype) in the population can be defined in a compact form. A context-free grammar is a quadruple (V, $\Sigma$, P, S), where V is a finite set of variables, $\Sigma$ is a finite set of terminal symbols, P is a finite set of rules or productions, and S the start symbol.

\begin{figure}[tb!]
\centering
\includegraphics[width=0.46\textwidth]{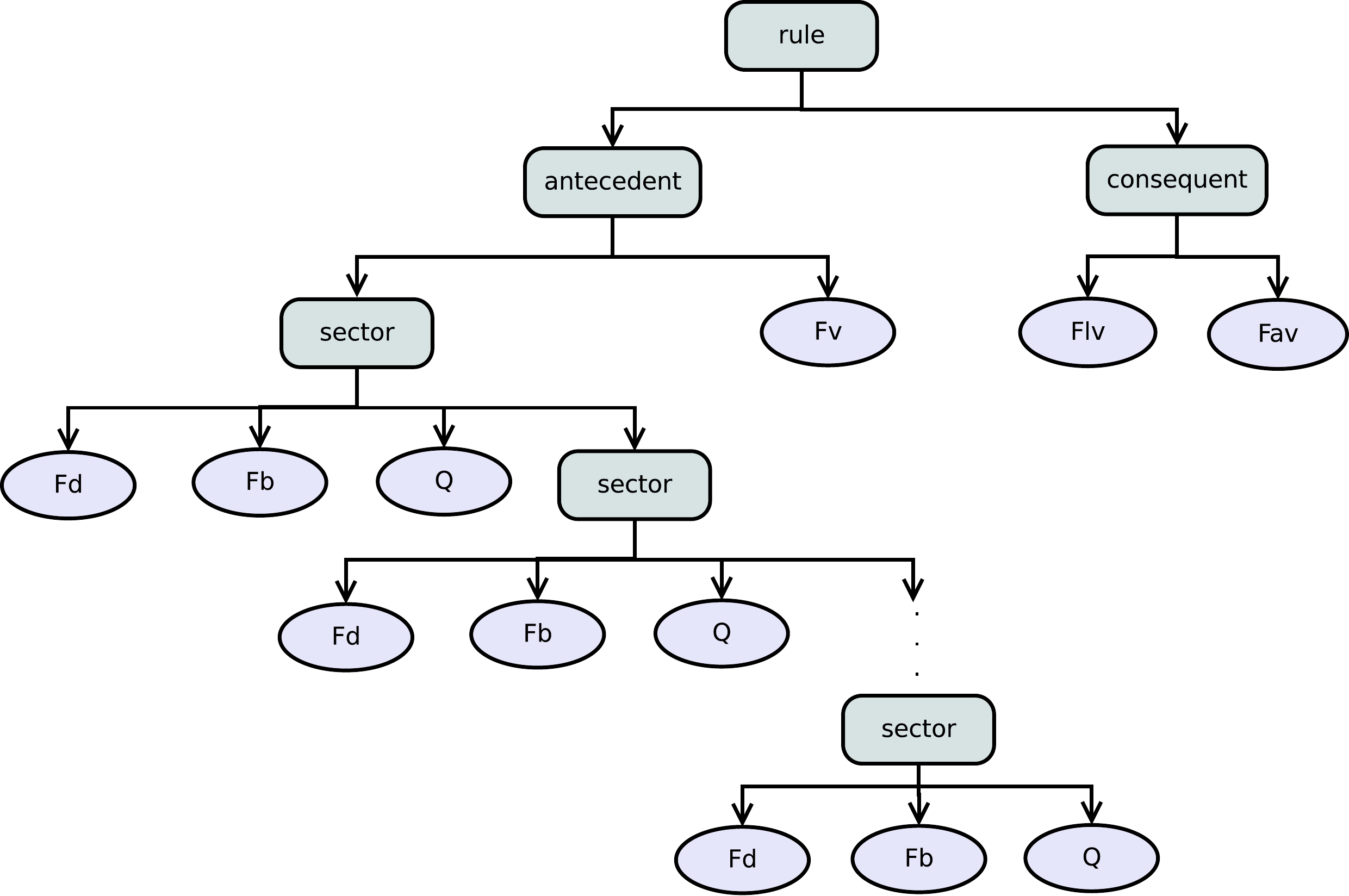}
\caption{\label{Fig:genotype}An individual representing a QFR that models the behavior of a robot.}
\end{figure}

The basic grammar is described in Fig. \ref{Fig:grammar}. As usual, different productions for the same variable are separated by symbol ``$\mid$''. Fig. \ref{Fig:genotype} represents a typical chromosome generated with this context-free grammar. Terminal symbols (leaves of the tree) are represented by ellipses, and variables as rectangles. There are two different types of antecedents:
\begin{itemize}
\item The sector antecedent. Consecutive beams are grouped into sectors in order to generate more general (high-level) variables (frontal distance, right distance, etc.). This type of antecedent is defined by the terminal symbols $F_d$, $F_b$ and $Q$: i) the linguistic label $F_d$ represents the measured distances ($HIGH$ in Fig. \ref{Fig:QFR}, prop. \ref{Eq:distProp_1}); ii) $F_b$ is the linguistic label that defines the sector, i.e., which beams belong to the sector ($FRONTAL\ SECTOR$ in Fig. \ref{Fig:QFR}, prop. \ref{Eq:distProp_1}); iii) $Q$ is the quantifier ($\textnormalit{part}$ in Fig. \ref{Fig:QFR}, prop. \ref{Eq:distProp_1}). 

\item The measured linear velocity of the antecedent is defined by the $F_v$ linguistic label.
\end{itemize}

Finally, $F_{lv}$ and $F_{av}$ are the linguistic labels of the linear and angular velocity control commands respectively, which are the consequents of the rule.

\begin{figure}[tb!]
\centering \setlength{\fboxrule}{0.2mm} \fbox{\parbox{0.47\textwidth}{
\setlength{\leftmargini}{0.5cm} 

\begin{itemize} 
\item V = $\lbrace$ rule, antecedent, consequent, sector $\rbrace$
\item $\Sigma$ = $\lbrace$ $F_{lv}$, $F_{av}$,
$F_v$, $F_d$, $F_b$, $Q$ $\rbrace$

\item $S$ = rule
\item P:

\begin{enumerate}
\item rule $\longrightarrow$ antecedent consequent
\item antecedent $\longrightarrow$ sector $F_v$ $\mid$ sector
\item consequent $\longrightarrow$ $F_{lv}$ $F_{av}$
\item sector $\longrightarrow$ $F_d$ $Q$ $F_b$ sector $\mid$ $F_d$ $Q$ $F_b$
\end{enumerate}

\end{itemize}
}}
\caption[]{\label{Fig:grammar}Basic context-free grammar for controllers in robotics.}
\end{figure}

\begin{figure}[tb!]
\centering
\includegraphics[width=0.5\textwidth]{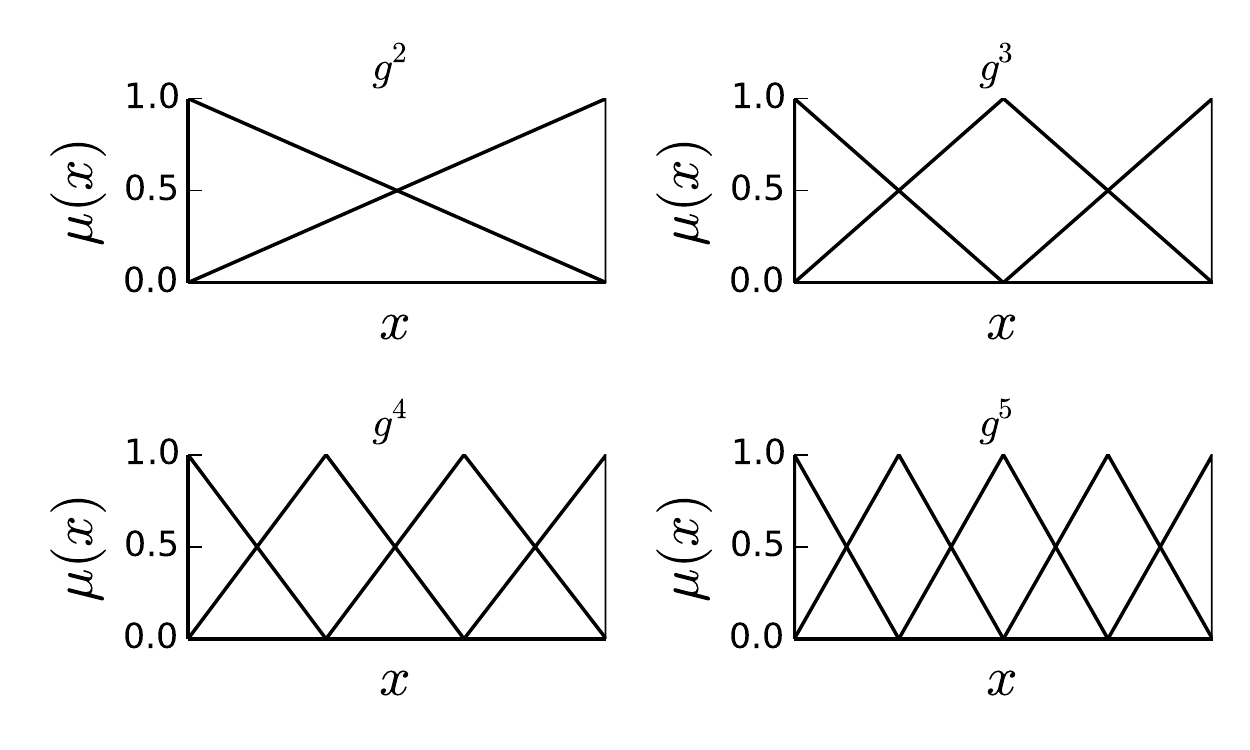}
\caption{\label{Fig:multipleg}Multiple granularity approach from $g_x^2$ to $g_x^5$.}
\end{figure}

The linguistic labels of the antecedent ($F_v$, $F_d$, $F_b$) are defined using a multiple granularity approach. The universe of discourse of a variable is divided into a different number of equally spaced labels for each granularity. Specifically, a granularity $g_{var}^i$ divides the variable $var$ in $i$ uniformly spaced labels, i.e., $A^i_{var} = \{ A^{i,\ 1}_{var}, ..., A^{i,\ i}_{var} \}$. 
Fig. \ref{Fig:multipleg} shows a partitioning of up to granularity five. On the other hand, the linguistic labels of the consequents ($F_{lv}$, $F_{av}$) are defined using a single granularity approach\footnote{Multiple granularity makes no sense if the labels are defined as singletons, which is the usual choice for the output variables in control applications.}.

\subsection{Initialization}
An individual (Fig. \ref{Fig:genotype}) is generated for each example in the training set. The consequent part ($F_{lv}$ and $F_{av}$) is initialized as $F_{var} = A^{g_{var},\ \beta}_{var}$ where $\beta = argmax_j\  \mu^{g_{var},\ j}_{var}\left(e^{l}\right)$, i.e., the label with the largest membership value for the example.

The initialization of the antecedent part of a rule requires obtaining the most similar linguistic label to a given fuzzy membership function (which is called mask label). As the maximum granularity of the linguistic labels in the antecedent part of a rule is not limited, the function \verb|maskToLabel| (Fig. \ref{Alg:mask}) is applied to obtain the most appropriate linguistic label. This function uses a similarity measure defined as \cite{scozzafava2009_fss}:
\setlength{\arraycolsep}{0.0em}
\begin{equation}\label{Eq:similarity}
similarity(F_\phi,\ F_\psi) = 1\ -\ \frac{\sum_{x \in X} |\mu_\phi(x)\ -\ \mu_\psi(x)|}{|X|}
\end{equation}
where $F_\phi$ and $F_\psi$ are the labels being compared and $X$ is a finite set of points $x$ uniformly distributed on the support of $\phi \cup \psi$.

The \verb|maskToLabel| function (Fig. \ref{Alg:mask}) receives a triangular membership function ($mask_{var}$) and searches for the label $A^{i,\ j}_{var}$ with the highest similarity (Eq. \ref{Eq:similarity}, line 6) with less or equal support (line 5), starting from $g^1_{var}$ (line 1).
\begin{figure}[tb!]
\begin{algorithmic}[1]
\REQUIRE $mask_{var}$
\STATE $i := g^1_{var}$
\STATE $result := \varnothing$
\LOOP
    \FORALL{$j \in [1, i]$}
	\IF{support($mask_{var}$) $\geq$ support($A^{i,\ j}_{var}$)}
	    \IF{similarity($mask_{var}$, $A^{i,\ j}_{var}$) $>$ \\similarity($mask_{var}$, $result$)}
		\STATE $result := A^{i,\ j}_{var}$
	    \ENDIF
	\ELSE
	    \STATE \textbf{break loop}
	\ENDIF
    \ENDFOR
    \STATE $i := i\ +\ 1$
\ENDLOOP
\RETURN $result$
\end{algorithmic}
\caption[]{\label{Alg:mask}Function that searches for the most similar label to $mask_{var}$.}
\end{figure}

\begin{figure}[tb!]
\centering
\subfigure[\emph{$mask_b$}\label{Fig:mask_beam}]{\includegraphics[width=0.23\textwidth]{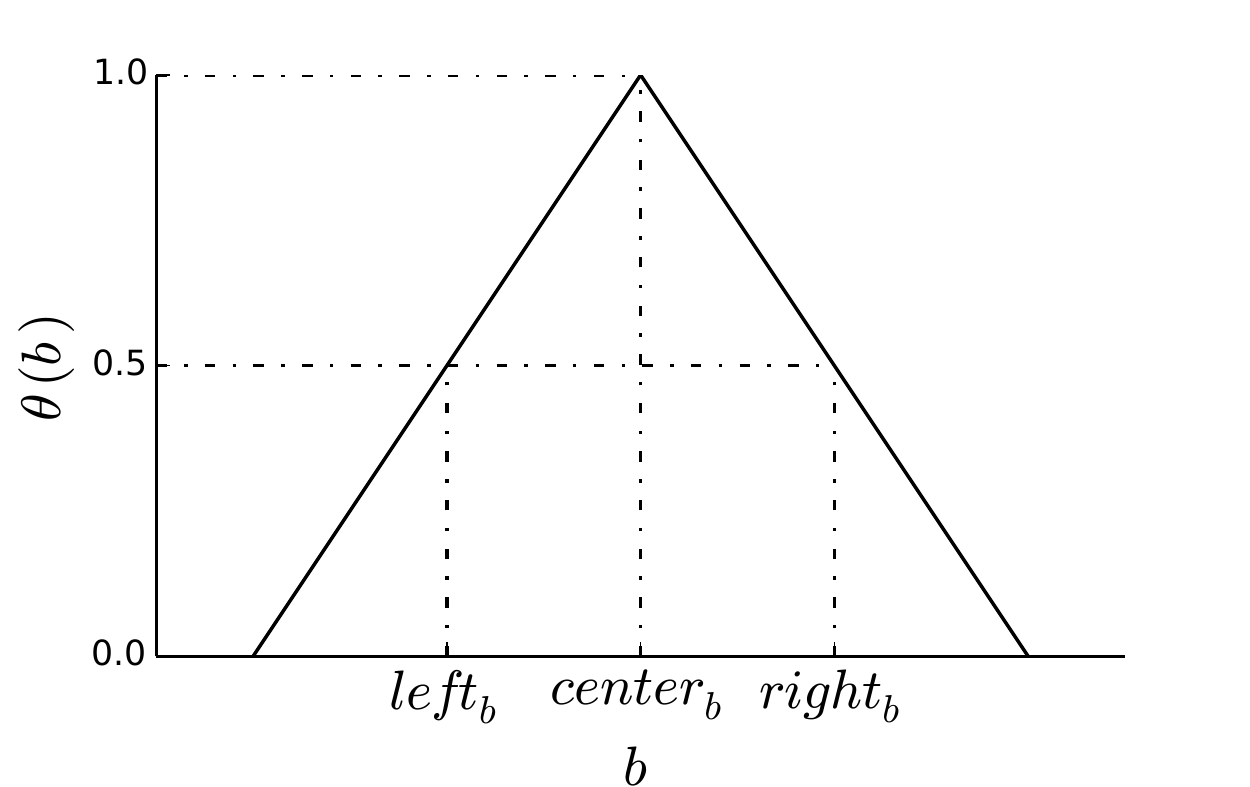}}
\subfigure[\emph{$mask_d$}\label{Fig:mask_distance}]{\includegraphics[width=0.23\textwidth]{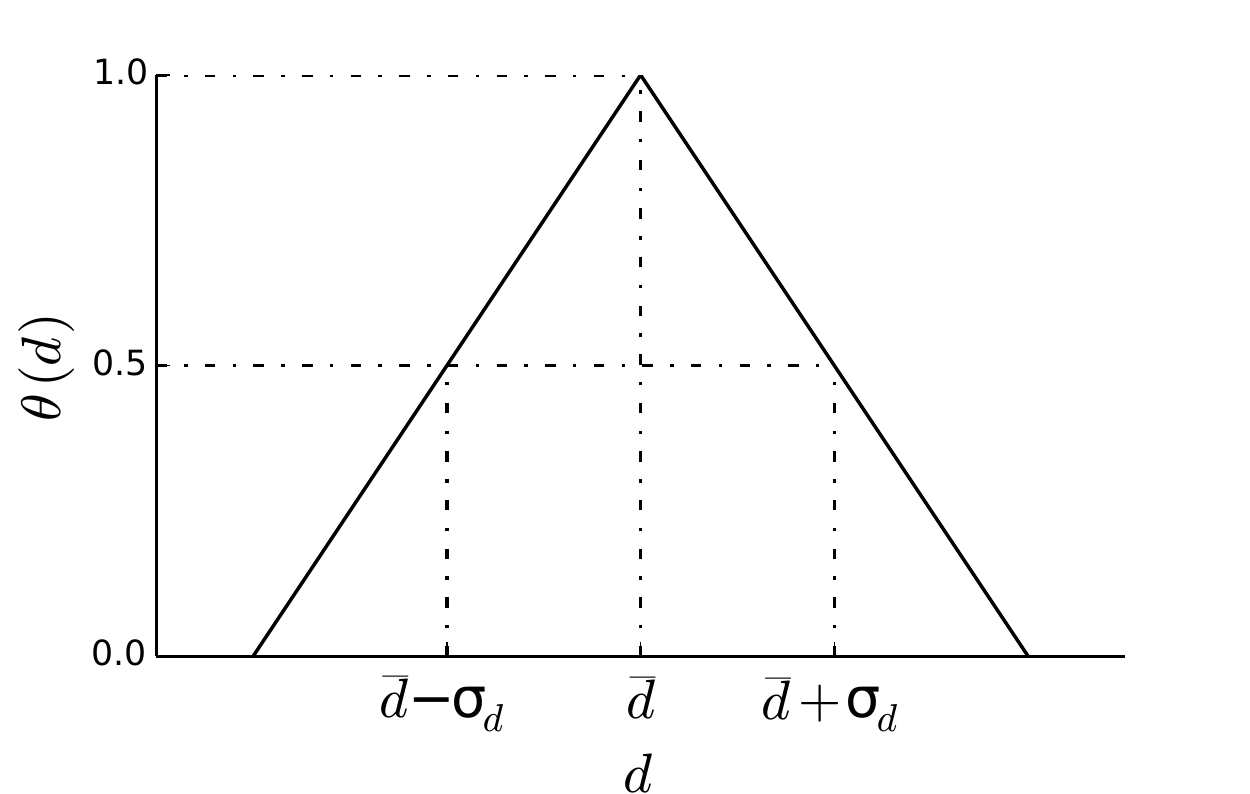}}
\caption{$mask_{var}$ representations for beam ($b$) and distance ($d$) variables.}

\end{figure}

\begin{figure}[tb!]
\centering
\includegraphics[width=0.25\textwidth]{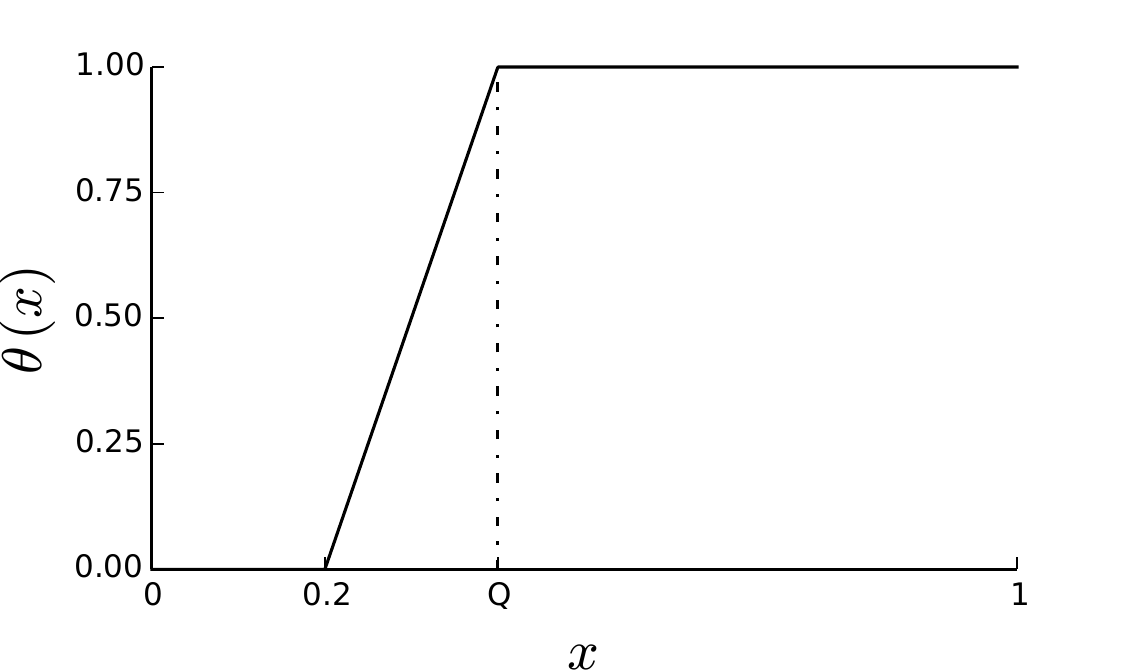}
\caption{Example of a definition of the quantified label $Q$.\label{Fig:Q}}
\end{figure}

For the initialization of the quantified propositions (sectors), the distances measured in the example are divided into groups of consecutive laser beams whose deviation does not exceed a certain threshold ($\sigma_{bd}$). Each group represents a sector that is going to be included in the individual. Afterwards, for each of the previously obtained sectors, the components ($F_b$, $F_d$ and $Q$) are calculated:
\begin{enumerate}
\item $F_b = maskToLabel(mask_{b})$, with $mask_{b} = (\textnormalit{left}_{b},\ \textnormalit{center}_{b},\ \textnormalit{right}_{b})$ where $\textnormalit{left}_{b}$ is the lower beam of the group, $\textnormalit{right}_{b}$ is the higher beam, $\textnormalit{center}_{b}$ is the middle beam and the following properties are satisfied: $\mu(\textnormalit{left}_{b}) = \mu(\textnormalit{right}_{b}) = 0.5$ and $\mu(\textnormalit{center}_{b}) = 1$ as shown in Fig. \ref{Fig:mask_beam}.
\item $F_d = maskToLabel(mask_{d})$, with $mask_{d} = (\bar{d}\ -\ \sigma_{d},\ \bar{d},\ \bar{d}\ +\ \sigma_{d})$ where $\bar{d}$ is the mean of the distances measured by the beams of the group, $\sigma_{d}$ is the standard deviation of these distances and the following properties are satisfied: $\mu(\bar{d}\ -\ \sigma_{d}) = \mu(\bar{d}\ +\ \sigma_{d}) = 0.5$ and $\mu(\bar{d}) = 1$ as shown in Fig. \ref{Fig:mask_distance}.
\item $Q$ (Fig. \ref{Fig:Q}) is calculated as the percentage of beams of the sector ($h \in F_b$) that fulfill $F_d$:
\begin{equation}
Q=\frac{\sum_{h \in F_b}min\left(\mu_{F_{d}}(d(h)),\ \mu_{F_{b}}(h)\right)}{\sum_{h \in F_b}\mu_{F_{b}}(h)}
\end{equation}

\end{enumerate}


Finally, the velocity antecedent $F_v$ is initialized as $F_v = A^{g_v^i,\ \beta}_v$ where $\beta = argmax_j\ \mu_v^{g_v^i,\ j}\ (e^l)$ and $g_v^i$ is the granularity that satisfies that two consecutive linguistic labels have a separation of $\sigma_v$, where $\sigma_v$ is a threshold of the velocity deviation.

\subsection{Evaluation}

The fitness of an individual of the population is calculated as follows. Firstly, it is necessary to estimate the probability that an example $e^l$ matches the output ($C_j$) associated to the $j$-th individual rule: 

\begin{equation}
\label{Eq:probEx}
P\left( C_{j}\ |\ e^l\right) =
\exp{\left(-\ \frac{\textnormalit{error}_j^l}{\textnormalit{ME}}\right)}
\end{equation}
where $\textnormalit{ME}$ is a parameter that defines the meaningful error and $\textnormalit{error}_j^l$ is the difference between output $C_j$ and the output codified in the example:

\begin{equation}
\label{Eq:error}
error_{j}^{l}=\sum_{k}\left(\frac{y_{k}^{l}\ -\ c_{j,\ k}}{max_{k}\ -\ min_{k}}\right)^{2}
\end{equation}
where $y_k^l$ is the value of the $k$-th output variable of example $e^l$, $c_{j,\ k}$ is the
output of the $k$-th output variable associated to individual $j$, and $max_{k}$ and $min_{k}$ are
the maximum and minimum values of output variable $k$. In regression
problems, there can be several consequents that are different from the one
codified in the example, but that produce small errors, i.e., that are very similar
to the desired output. Thus, $P\left( C_j\ |\ e^l\right)$ can be interpreted as
a normal distribution with covariance $ME$, and $error_{j}^{l}$ is the square of
the difference between the mean (output codified in the example) and the output
value proposed in the rule codified by the individual.

In an IRL approach, $C_j = C_{R_j}$, i.e., the output coded in individual $j$ is the output associated to rule $j$. The fitness of an individual in the population is calculated as the combination of two values. On one hand, the accuracy with which the individual covers the examples, called confidence. On the other hand, the ability of generalization of the rule, called support. The confidence can be defined as:
\begin{equation}
\textnormalit{confidence}=\frac{\rho_{u}}{\sum_l DOF_{j}(e_u^l)}
\label{Eq:confidence}
\end{equation}
where $DOF_{j}(e_u^l)$ is the degree of fulfillment of $e_u^l$ for rule $j$, and $e^l_u \in uncov_{ex}$, where $uncov_{ex}$ is defined as:
\begin{equation}
uncov_{ex} = \{ e^l  \ :\ DOF_{\textnormalit{KB}_{\textnormalit{cur}}}(e^l) < DOF_{min}\}
\end{equation}
i.e., the set of examples that are covered with a degree of fulfillment below $DOF_{min}$ by the current final knowledge base ($KB_{cur}$) (line \ref{Eq:finalkb}, Fig. \ref{Alg:IQFRL}),  and $\rho_{u}$ can be defined as: 
\begin{equation}
\label{Eq:accuracy_u}
\begin{aligned}
\rho_{u}=\sum_l DOF_{j}(e_u^l)\ :\ P\left( C_{j}\ |\ e_u^l\right) > P_{min}\\
and\ DOF_{j}(e_u^l) > DOF_{min}
\end{aligned}
\end{equation}
where $P_{min}$ is the minimum admissible accuracy. Therefore, the higher the accuracy over the examples covered by the rule (and not covered yet by the current knowledge base), the higher the confidence. Support is calculated as: 

\begin{equation}
support=\frac{\rho_{u}}{\# uncov_{ex}}
\end{equation}
Thus, support measures the percentage of examples that are covered with accuracy, related to the total number of uncovered examples. Finally, $fitness$ is defined as a linear combination of both values: 
\begin{equation}
\label{Eq:fitness}
\textnormalit{fitness}=\alpha_f \cdot \textnormalit{confidence} + (1 - \alpha_f) \cdot support
\end{equation}
which represents the strength of an individual over the set of examples in $uncov_{ex}$. $\alpha_f \in [0,\ 1]$ is a parameter that codifies the trade-off between accuracy and generalization of the rule.

\subsection{Crossover}

The matching of the pairs of individuals that are going to be crossed is implemented following a probability distribution defined as:
\begin{equation}
P_{close}\left( \alpha,\ \beta \right) = 1\ -\ \frac{\sum^{N_c}_{k=1}(\frac{c_{\alpha,\ k}\ -\ c_{\beta,\ k}}{max_{k}\ -\ min_{k}})^{2}}{N_c}
\end{equation}
where $c_{\alpha,\ k}$ ($c_{\beta,\ k}$) is the value of the $k$-th output variable of individual $\alpha$ ($\beta$), and $N_c$ is the number of consequents. With this probability distribution, the algorithm selects with higher probability mates that have similar consequents. The objective is to extract information on which propositions of the antecedent part of the rules are important, and which are not. Crossover has been designed to generate more general individuals, as the initialization of the population produces very specific rules. The crossover operator generates two offsprings: 
\begin{eqnarray}
\textnormalit{offspring}_1 = crossover(ind_i,\ ind_j)\nonumber\\
\textnormalit{offspring}_2 = crossover(ind_j,\ ind_i)
\end{eqnarray}

This operator modifies a single proposition in antecedent part of the rule. As individuals have a variable number of antecedents, the total number of propositions can be different for two individuals. Moreover, the propositions can be defined using different granularities. Therefore, the first step is to select the propositions (one for each individual) to be crossed between both individuals (Fig. \ref{Alg:crossover}) as follows:

\begin{enumerate}
\item Get the most specific granularity of the sectors of the individuals to cross ($g_b^{max}$). Then, an antecedent $m \in [1,\ N_a]$ is selected, where $N_a$ is $g_b^{max}$ plus one, due to the velocity proposition.
\item Check the existence of this antecedent in both individuals, according to the following criteria:
    \begin{enumerate}
    \item If the antecedent $m$ is a sector, then calculate for each proposition of each individual the similarity between the definition of the sector for the proposition and the linguistic label that defines sector $m$. Finally, select for each individual the proposition with the highest similarity.
    
    \item If the antecedent $m$ is the velocity, then the corresponding proposition is $F_v$ (in case it exists).
    \end{enumerate}
\end{enumerate}

\begin{figure}[tb!]
\begin{algorithmic}[1]
\REQUIRE ${ind_{\alpha},\ ind_{\beta}}$
\STATE $a_{\alpha} = a_{\beta} = \varnothing$
\STATE $N_a = g_b^{max}\ +\ 1$
\REPEAT
    \STATE $m = random \in [1,\ N_a]$
    \IF{$m$ is a $sector$}
	\STATE $a_{\alpha} = argmax_r\ similarity(F_{b,\ r},\ A^{g^{max}_{b}, m}_b) \geq 0 : \forall r \in ind_{\alpha}$
	\STATE $a_{\beta} = argmax_r\ similarity(F_{b,\ r},\ A^{g^{max}_{b}, m}_b) \geq 0 : \forall r \in ind_{\beta}$
    \ELSE
	\STATE $a_{\alpha} = F_v \in ind_{\alpha}$
	\STATE $a_{\beta} = F_v \in ind_{\beta}$
    \ENDIF
\UNTIL{$(a_{\alpha} \neq \varnothing) \vee (a_{\beta} \neq \varnothing)$}
\end{algorithmic}
\caption[]{\label{Alg:crossover}Selection of antecedents for crossover.}
\end{figure}

Once the propositions to be crossed have been selected, an operation must be picked depending on the existence of the antecedent in both parents (table \ref{Tab:crossover}): 
\begin{itemize}
\item If the proposition does not exist in the first individual but exists in the second one, then the proposition of the second individual is copied to the first one, as this proposition could be meaningful.
\item If the situation is the opposite to the previous one, then the proposition of the first individual is deleted, as it might be not important.
\item If the proposition exists in both individuals, then both propositions are combined in order to obtain a proposition that generalizes  both antecedents.
\end{itemize}

\begin{table}[tb!]
\centering
\footnotesize
\caption{\label{Tab:crossover}Crossover operations}
\begin{tabular}{|c|c|c|}
\hline
Individual 1 & Individual 2 & Action\\
\hline
\hline
no & yes & copy proposition from individual 2 to 1\\
yes & no & delete proposition in individual 1\\
yes & yes & combine propositions\\
\hline
\end{tabular}
\end{table}

\begin{figure}[tb!]
\centering
\includegraphics[width=0.5\textwidth]{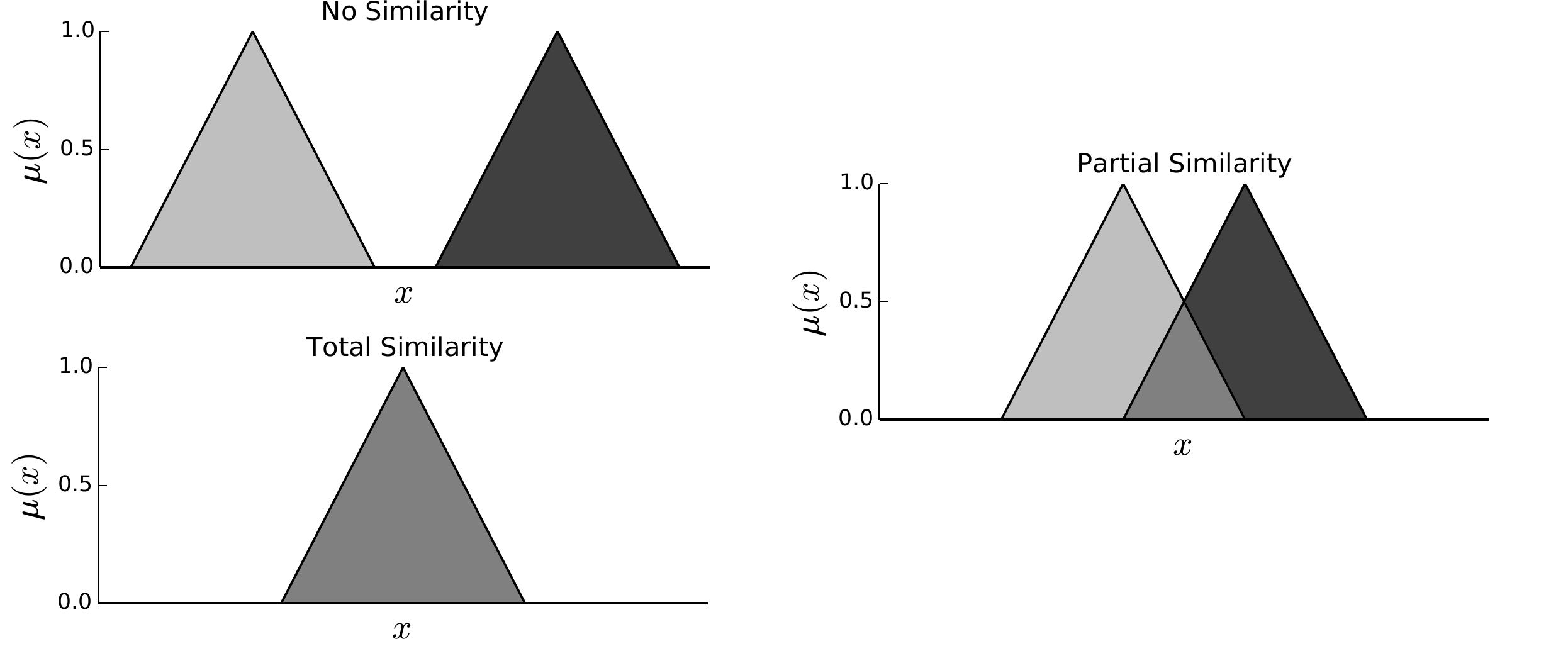}
\caption{\label{Fig:cross} Different possibilities of similarity for the labels of equal proposition of two individuals used in the crossover operator.}
\end{figure}

In this last case, the combination of propositions is done by taking into account the degree of similarity (Eq. \ref{Eq:similarity}) between them (Fig. \ref{Fig:cross}). If the proposition is of type sector, the similarity takes into account both $F_b$ and $F_d$ labels. Only when both similarities are partial, the propositions are merged: 
\begin{itemize}
\item If there is no similarity, then the propositions correspond to different situations. For example, \textit{``the distance is high in part of the frontal sector''} and \textit{``the distance is low in part of the frontal sector''}. This means that the proposition of the first individual might not contain meaningful information and it could be deleted to generalize the rule. For example, both individuals have the proposition \textit{``the distance is high in part of the frontal sector''}.
\item If the similarity is total, then, in order to obtain a new individual with different antecedents, the proposition is eliminated.
\item Finally, if the similarity is partial, then the propositions are merged in order to obtain a new one that combines the information provided by the two original propositions. For example, \textit{``the distance is high in part of the frontal sector``} and \textit{``the distance is medium-high in part of the frontal sector``}. Therefore, the individual is generalized. The merge action is defined as the process of finding the label with the highest possible granularity that has some similarity with the labels of both original propositions. This is done for both $F_b$ and $F_d$ labels. $Q$ is calculated as the minimum $Q$ of both individuals.
\end{itemize}

\subsection{Mutation}

\begin{figure}[tb!]
\centering
\includegraphics[width=0.45\textwidth]{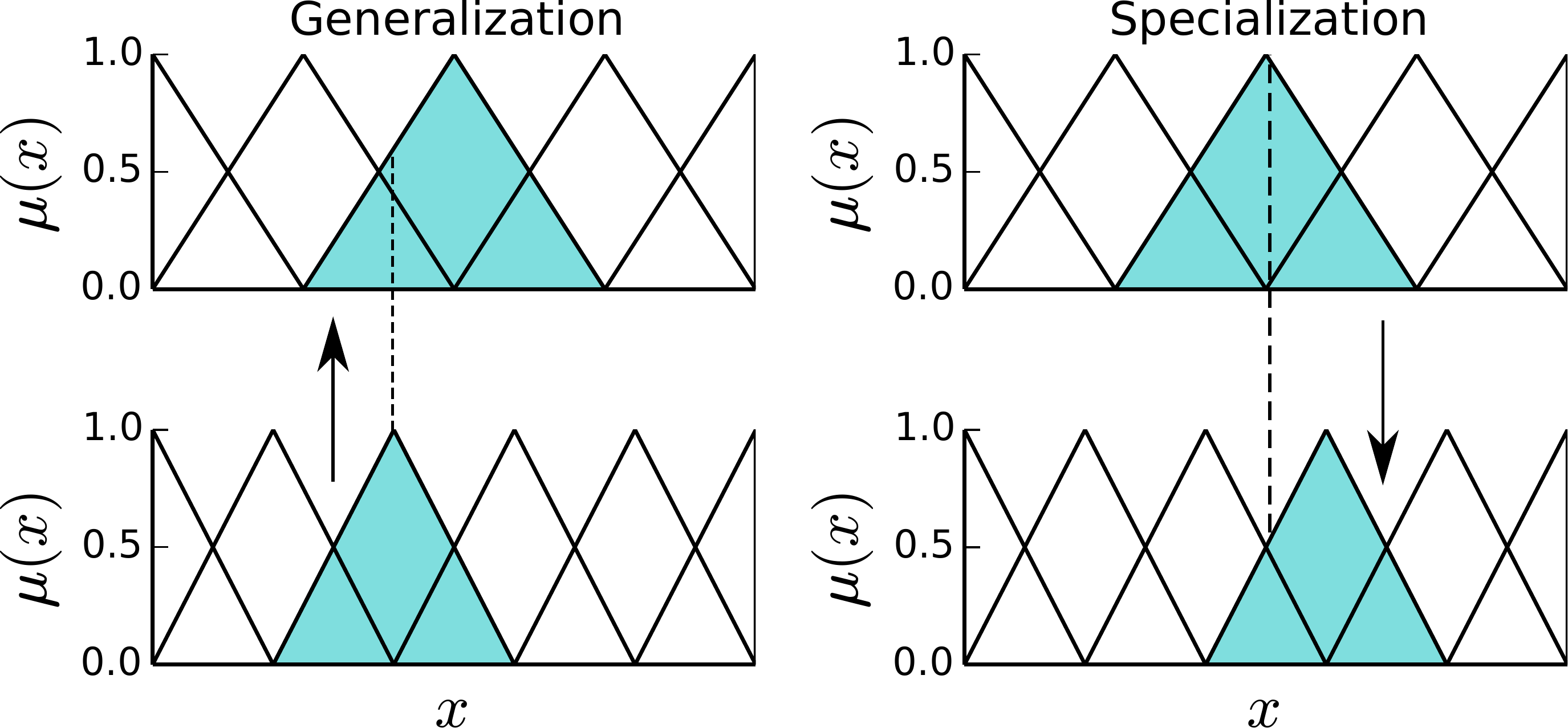}
\caption{\label{Fig:mutation}The strategies used for mutation for variables $d$, $b$ and $v$.}
\end{figure}

If crossover is not performed, both individuals are mutated. Mutation implements two different strategies (Fig. \ref{Fig:mutation}): generalize or specialize a rule. The higher the value of confidence (Eq. \ref{Eq:confidence}), the higher the probability to generalize the rule by mutation. This occurs with rules that cover their examples with high accuracy and that could be modified to cover other examples. On the contrary, when the confidence of the individual is low, this means that it is covering some of its examples with a low performance. In order to improve the rule some of the examples that are currently covered should be discarded in order to get a more specific rule.

For generalization, the following steps are performed: 
\begin{enumerate}
\item Select an example $e^{sel} \in uncov^j_{ex}$, where $uncov^j_{ex} = \{e^l_u : DOF_{j}(e^l_u) < DOF_{min}\}$, i.e. the set of examples that belong to $uncov_{ex}$ and are not covered by individual $j$. The example is selected with a probability distribution given by $P\left( C_{j}\ |\ e^l_u\right)$ (Eq. \ref{Eq:probEx}). The higher the similarity between the output of the example and the consequent of rule $j$, the higher the probability of being selected. 
\item The individual is modified in order to cover $e^{sel}$. Therefore, all the propositions that are not covering the example (those with $\mu_{prop}\left(e^{sel}\right) < DOF_{min}$) are selected for mutation.
\begin{enumerate}
\item For sector propositions (Eq. \ref{Eq:distProp_1}), there are three different ways in which the proposition can be modified: $F_d$, $F_b$, and $Q$. The modification is selected among the three possibilities, with a probability proportional to the $\mu_{prop}\left(e^{sel}\right)$ value after applying each one. \begin{enumerate}
\item $F_d$ and $F_b$ are generalized choosing the most similar label in the adjacent partition with lower granularity. The process is repeated until $\mu_{prop}\left(e^{sel}\right) \geq DOF_{min}$. 
\item On the other hand, $Q$ is decreased until $\mu_{prop}\left(e^{sel}\right) \geq DOF_{min}$.
\end{enumerate}
\item For velocity propositions (Eq. \ref{Eq:velProp}),  generalization is done choosing the most similar label in the adjacent partition with lower granularity until $\mu_{prop}\left(e^{sel}\right) > DOF_{min}$.
\end{enumerate}
\end{enumerate}

For specialization, the process is equivalent: 
\begin{enumerate}
\item Select an example $e^{sel} \in cov^j_{ex}$, where $cov^j_{ex} = \{e^l_u : DOF_{j}(e^l_u) > DOF_{min}\}$, i.e. the set of examples that belong to $uncov_{ex}$ and are covered by individual $j$. The example is selected with a probability distribution that is inversely proportional to $P\left( C_{j}\ |\ e^l_u\right)$ (Eq. \ref{Eq:probEx}). The higher the similarity between the output of the example and the consequent of rule $j$, the lower the probability of being selected. 
\item Only one proposition needs to be modified to specialize the individual. This proposition is selected randomly.
\begin{enumerate}
\item For sector propositions there are, again, three different ways in which the proposition can be modified: $F_d$, $F_b$, and $Q$. The modification is selected among these three possibilities, with a probability that is inversely proportional to the $\mu_{prop}\left(e^{sel}\right)$ value after applying each one. 
\begin{enumerate}
\item $F_d$ and $F_b$ are specialized, choosing the most similar label in the adjacent partition with higher granularity. The process is repeated until $\mu_{prop}\left(e^{sel}\right) < DOF_{min}$. 
\item On the other hand, $Q$ is increased until $\mu_{prop}\left(e^{sel}\right) < DOF_{min}$. 
\end{enumerate}
\item For velocity propositions, specialization is done by choosing the most similar label in the adjacent partition with higher granularity until $\mu_{prop}\left(e^{sel}\right) < DOF_{min}$.
\end{enumerate}
\end{enumerate}

\begin{figure}[tb!]
\centering
\includegraphics[width=0.35\textwidth]{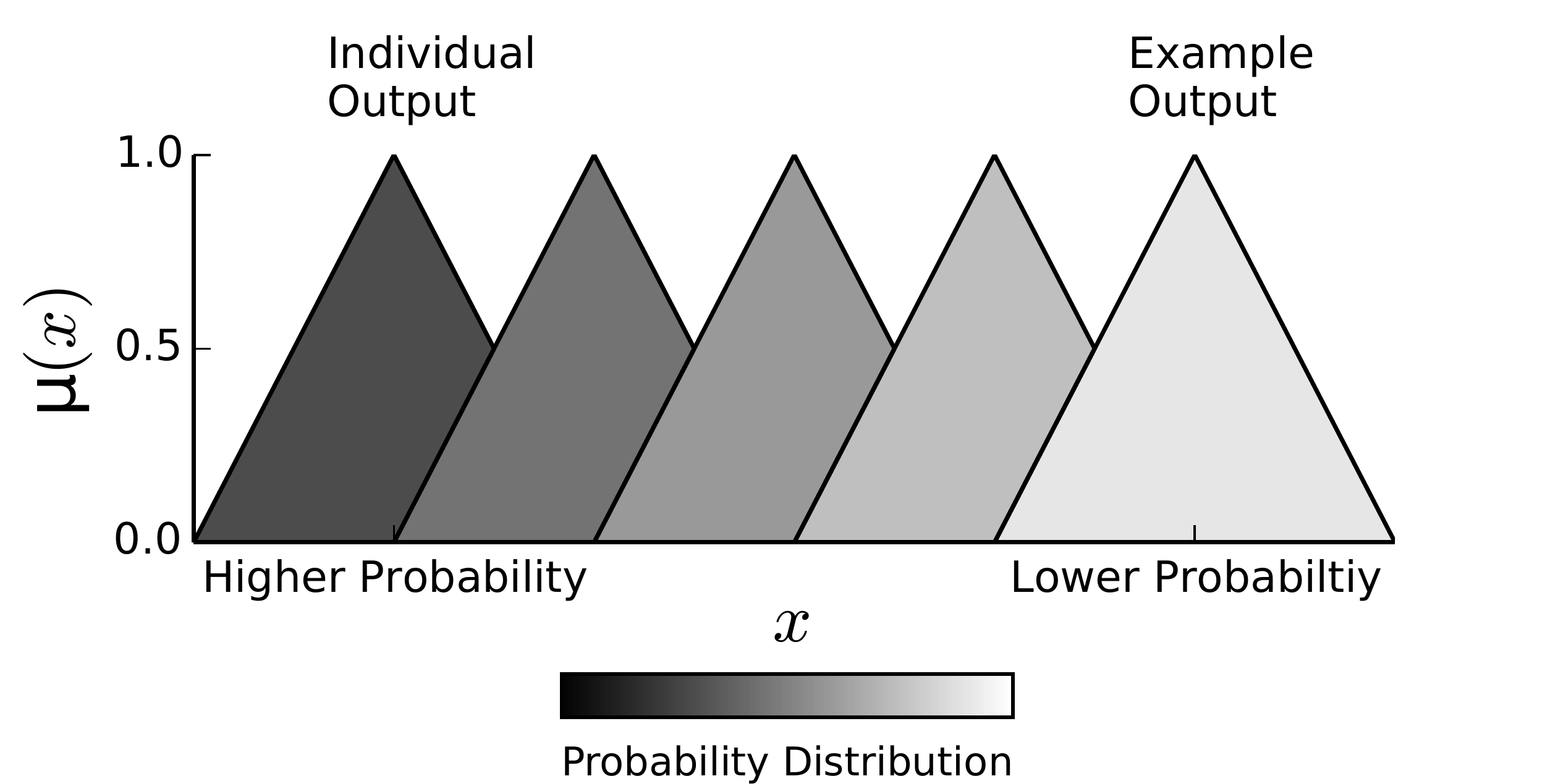}
\caption{\label{Fig:consequent}Probability distribution example for consequent mutation. Labels closest to the individual output have higher probability to be selected.}
\end{figure}

Finally, once the antecedent is mutated, the consequent also mutates. Again, this mutation requires the selection of an example. If generalization was selected for the mutation of the antecedent, then the example will be $e^{sel}$. On the other hand, for specialization an example is randomly selected from those currently in $cov^j_{ex}$. For each variable in the consequent part of the rule, the label of the individual is modified selecting a label following a probability distribution (Fig. \ref{Fig:consequent}):
\begin{equation}
 P\left(A_{var}^{g_{var},\ \gamma}\ |\ A_{var}^{g_{var},\ \alpha},\ A_{var}^{g_{var},\ \beta}\right) = 1\ -\ \frac{|\alpha\ -\ \gamma|}{|\alpha\ -\ \beta|\ +\ 1}
\end{equation}
where $A_{var}^{g_{var},\ \alpha}$ is the label of each of the consequents of the individual, $A_{var}^{g_{var},\ \beta}$ is the label with the largest membership value for $e^{sel}$ and $A_{var}^{g_{var},\ \gamma}$ is a label between them. Thus, the labels closer to the label of the individual have a higher probability to be selected, while the labels closer to the example label have a lower one.

\subsection{Selection and replacement}

Selection has been implemented following the binary tournament strategy. Replacement follows an steady-state approach. The new individuals and those of the previous population are joined, and the best $pop_{\textnormalit{max}}$ individuals are
selected for the next population. 

\subsection{Epoch loop}
An epoch is a set of iterations at the end of which a new rule is added to $\textnormalit{KB}_{\textnormalit{cur}}$. The stopping criterion of each epoch (inner loop in Fig. \ref{Alg:IQFRL}) is the number of iterations, but this limit varies according to the following criteria: once the number of iterations ($it$) reaches $it_{min}$, the algorithm stops if there are $it_{check}$ consecutive iterations (counted by $equal_{ind}$) with no change in the best individual ($best_{ind}$). If the number of iterations reaches the maximum ($it_{max}$), then the algorithm stops regardless of the previous condition. 

When the epoch ends, the rule defined in $best_{ind}$ is added to $\textnormalit{KB}_{\textnormalit{cur}}$. Moreover, the examples that are covered with accuracy (according to the criterion in Eq. \ref{Eq:accuracy_u}) are marked as covered by the algorithm (line \ref{Eq:remained}, Fig. \ref{Alg:IQFRL}). Finally, the algorithm stops when there are no uncovered examples.

\subsection{Rule subset selection\label{Sec:ruleSelection}}
 
After the end of the iterative part of the algorithm, the performance of the obtained rule base can be improved selecting a subset of rules with better cooperation among them. The rule selection algorithm described in \cite{mucientes2010_pr} has been used. The rule selection process has the following steps:

\begin{enumerate}
 \item Generate $\# R_{gp}$ rule bases, where $\# R_{gp}$ is the number of rules of the population obtained by the IQFRL algorithm ($RB_{gp}$) Each rule base is coded as: $RB_i = r^i_1 \cdots r^i_{\# R_{gp}}$, with:
\begin{equation}
 r^i_j = \begin{cases}
                 0, & if\ j>i\\
		 1, & if\ j\leq i
                 \end{cases}
\end{equation}
where $r^i_j$ indicates if the $j$-th rule of $RB_{gp}$ is included ($r^i_j = 1$) or not ($r^i_j = 0$) in $RB_i$. With this codification, $RB_i$ will contain the best $i$ rules of $RB_{gp}$, as these rules have been ranked in decreasing order of their individual fitness. Notice that $RB_{\# R_{gp}}$ is $RB_{gp}$
\item Evaluate all the rule bases, and select the best one, $RB_{sel}$.
\item Execute a local search on $RB_{sel}$ to obtain the best rule set, $RB_{best}$.
\end{enumerate}
The last step was implemented with the iterated local search (ILS) algorithm \cite{Lourenco03}. 

threshold (maxRestarts).

\section{Results\label{Sec:results}}
\subsection{Experimental setup}
\begin{figure}[tb!]
\centering
\includegraphics[width=0.3\textwidth]{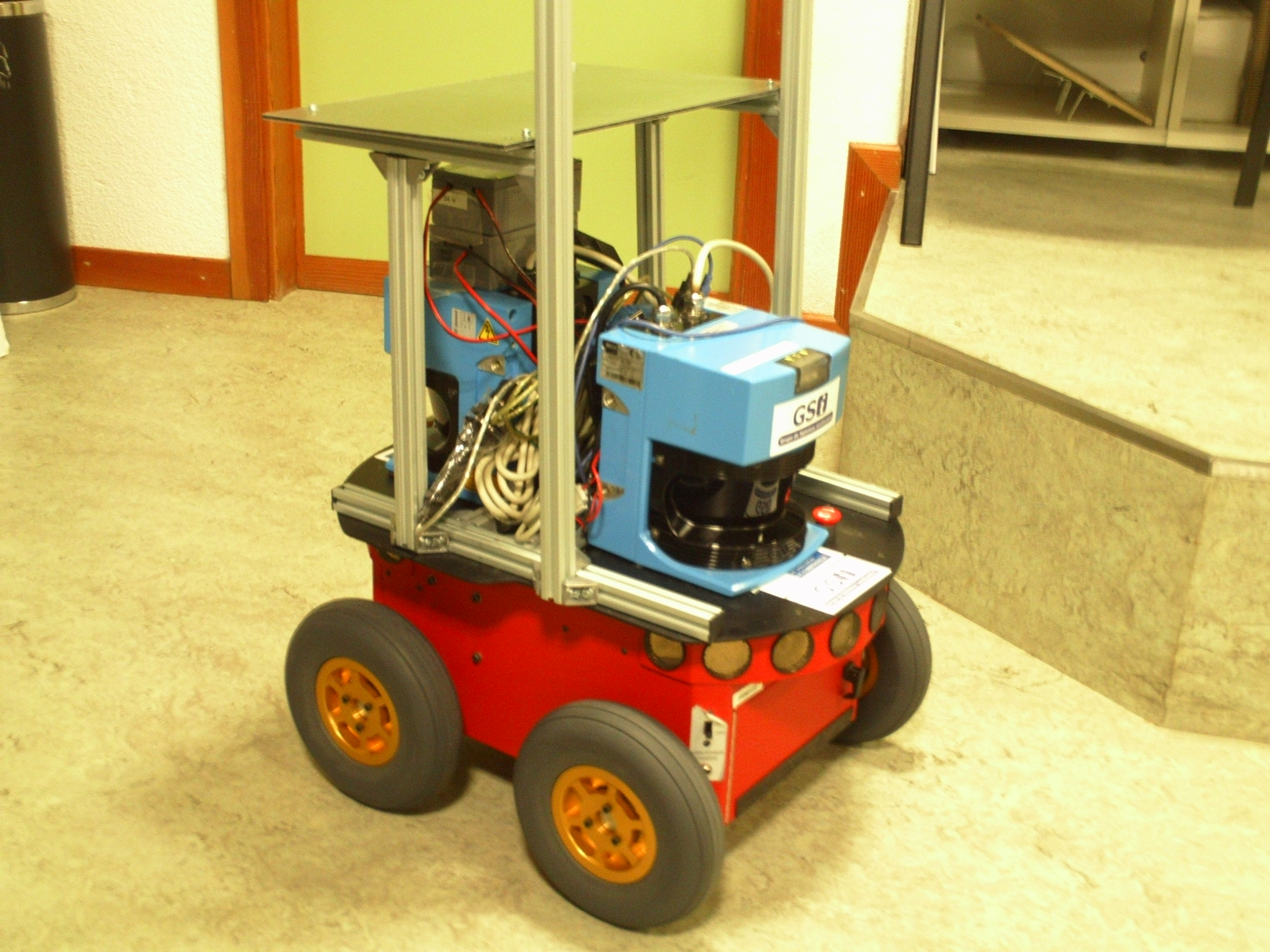}
\caption{\label{Fig:3at}\textit{Pioneer 3-AT} robot equipped with two laser range scanners.}
\end{figure}

The proposed algorithm has been validated with the well-known in mobile robotics wall-following behavior. The main objectives of a controller for this behavior are: to keep a suitable distance between the robot and the wall, to move at the highest possible velocity, and to implement smooth control actions. The Player/Stage robot software \cite{PlayerStage} has been used for the tests on the simulated environments and also for the connection with the real robot \textit{Pioneer 3-AT} (Fig. \ref{Fig:3at}). This real robot was equipped with two laser range scanners with an amplitude of $180^{\circ}$ and a precision of $0.5^{\circ}$ (i.e. 361 measurements for each laser scan). Without loss of generality, all the examples and tests here described were made with the robot following the wall at its right.

\begin{figure}[tb!]
\centering
\includegraphics[width=0.2\textwidth]{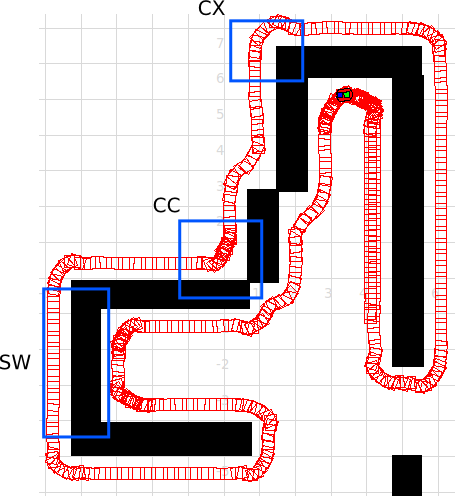}
\caption{\label{Fig:situations}The three different situations for the wall-following behavior.}
\end{figure}

The examples that have been used for learning were generated for three different situations (Fig. \ref{Fig:situations}) that have been identified by an expert:
\begin{enumerate}
\item Convex corner: it is characterized by the existence of a gap in the wall (like an open door) (labeled CX in Fig. \ref{Fig:situations}).
\item Concave corner: it is a situation in which the robot finds a wall in front of it (labeled CC in Fig. \ref{Fig:situations}).
\item Straight wall: any other situation (labeled SW in Fig. \ref{Fig:situations}).
\end{enumerate}

For each of the above situations, the robot was placed in different positions and the associated control order was the one that minimized the error. Therefore, each example consists of 722 distances (one for each laser beam), the current linear velocity of the robot, and the control commands (linear and angular velocity). The expert always tried to follow the wall at, approximately, 50 cm and the maximum values for the linear and angular velocities were 50 cm/s and $45^{o}s^{-1}$ respectively. 572 training examples were generated for the straight wall situation, 540 for the convex corner and 594 for the concave corner.

The IQFRL algorithm was used to learn a different controller for each of the three situations. In order to decide which knowledge base should be used at each time instant, the classification version of IQFRL (IQFRL-C, see \ref{Sec:class}) was used. In this way, IQFRL learning could be tested with three completely different controllers.

In order to analyze the performance of the proposed learning algorithm, several tests were done in 15 simulated environments and two real ones. Table \ref{Tab:enviroments} shows some of the characteristics of the environments: the dimensions of the environment, the path length, the number of concave (\#CC) and convex (\#CX) corners, and the number of times that the robot has to cross a door (\#doors). The action of crossing a door represents a high difficulty as the robot has to negotiate a convex corner with a very close wall in front of it.

\begin{table}[tb!]
\footnotesize
\centering
\caption{\label{Tab:enviroments}Characteristics of the test environments}
\begin{tabular}{|c|c|c|c|c|c|}
\hline
Environment & Dim. ($m \times m$) & Length (m) & \#CC & \#CX & \#doors\\
\hline
\hline
home & $8 \times 10$ & $20$ & $8$ & $3$ & $1$\\
gfs\_b & $14 \times 10$ & $43$ & $10$ & $6$ & $0$\\
dec & $19 \times 12$ & $53$ & $8$ & $4$ & $0$\\
domus & $26 \times 16$ & $60$ & $9$ & $6$ & $3$\\
citius & $16 \times 10$ & $63$ & $12$ & $6$ & $2$\\
raid\_a & $16 \times 16$ & $66$ & $16$ & $12$ & $0$\\
wsc8a & $15 \times 15$ & $70$ & $4$ & $7$ & $1$\\
home\_b & $18 \times 11$ & $76$ & $17$ & $6$ & $2$\\
raid\_b & $20 \times 10$ & $86$ & $12$ & $10$ & $2$\\
rooms & $19 \times 19$ & $86$ & $12$ & $6$ & $4$\\
flower & $22 \times 20$ & $98$ & $9$ & $6$ & $1$\\
office & $26 \times 26$ & $146$ & $23$ & $10$ & $8$\\
autolab & $26 \times 28$ & $154$ & $21$ & $11$ & $10$\\
maze & $18 \times 18$ & $205$ & $13$ & $9$ & $0$\\
hospital & $74 \times 45$ & $1046$ & $98$ & $69$ & $43$\\
real env 1 & $9 \times 8$ & $20$ & $7$ & $3$ & $0$\\
real env 2 & $10 \times 5$ & $26$ & $7$ & $3$ & $0$\\
\hline
\end{tabular}
\end{table}

\begin{figure*}[htb!]
\centering
\begin{tabular}{>{\centering\arraybackslash}m{0.28\textwidth}>{\centering\arraybackslash}m{0.28\textwidth}>{\centering\arraybackslash}m{0.28\textwidth}}
\subfigure[\emph{home}\label{Fig:home}]{\includegraphics[scale=0.22]{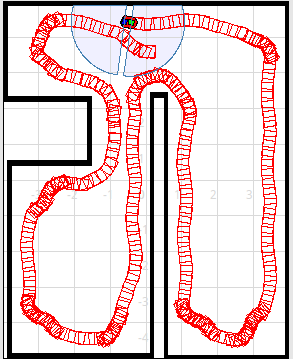}} &
\subfigure[\emph{gfs$\_$b}\label{Fig:gfs_b}]{\includegraphics[scale=0.22]{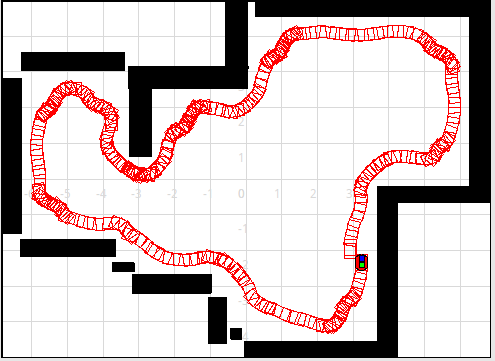}} &
\subfigure[\emph{dec}\label{Fig:dec}]{\includegraphics[scale=0.22]{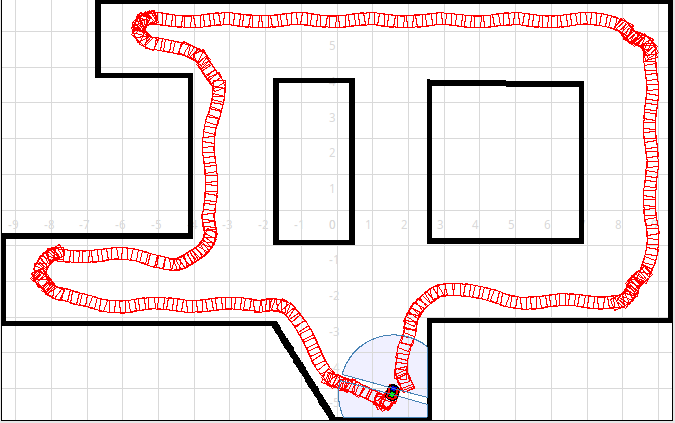}}\\
\subfigure[\emph{domus}\label{Fig:domus}]{\includegraphics[scale=0.22]{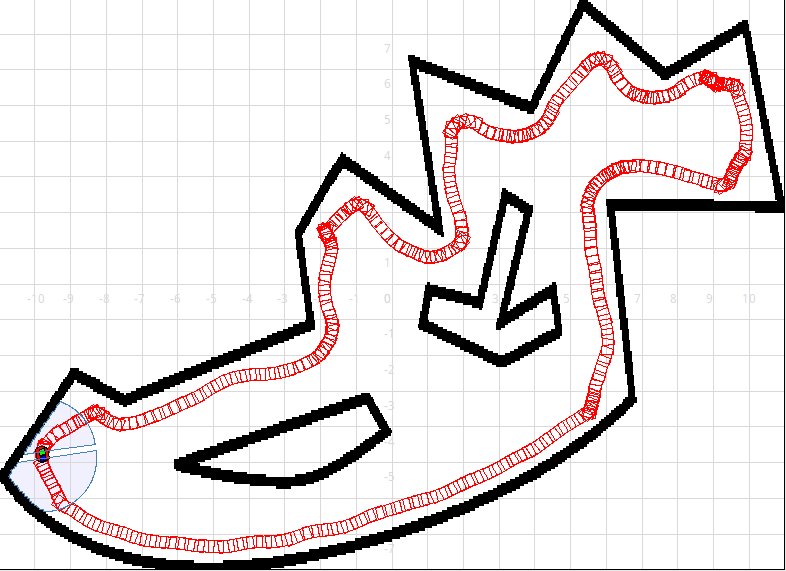}} &
\subfigure[\emph{citius}\label{Fig:citius}]{\includegraphics[scale=0.25]{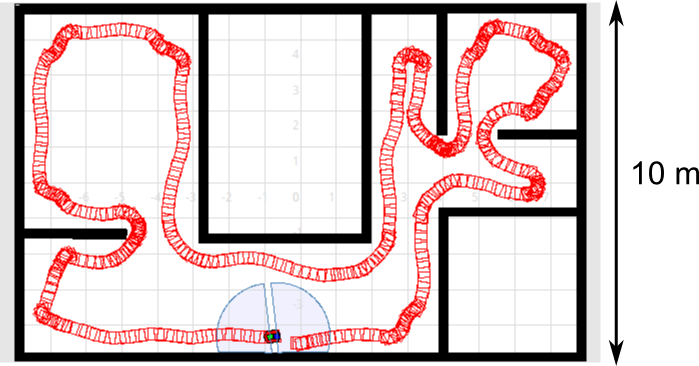}} & 
\subfigure[\emph{raid\_a}\label{Fig:raid_a}]{\includegraphics[scale=0.22]{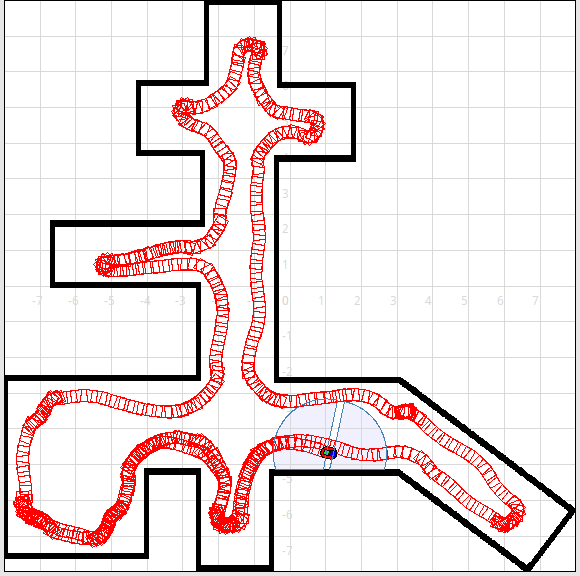}}\\
\subfigure[\emph{wsc8a}\label{Fig:wsc8a}]{\includegraphics[scale=0.22]{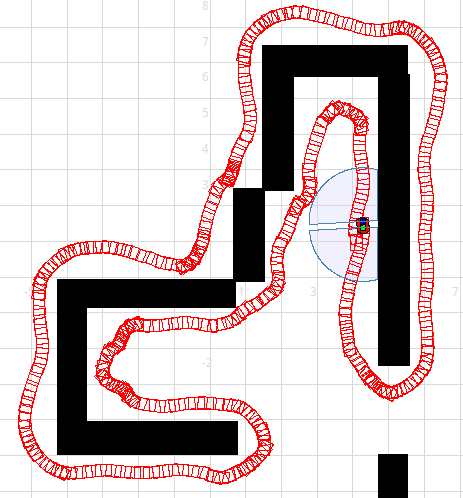}} &
\subfigure[\emph{home\_b}\label{Fig:home_b}]{\includegraphics[scale=0.25]{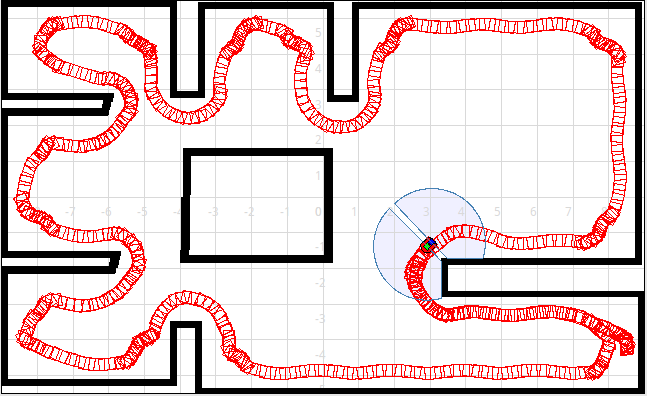}} &
\subfigure[\emph{raid\_b}\label{Fig:raid_b}]{\includegraphics[scale=0.22]{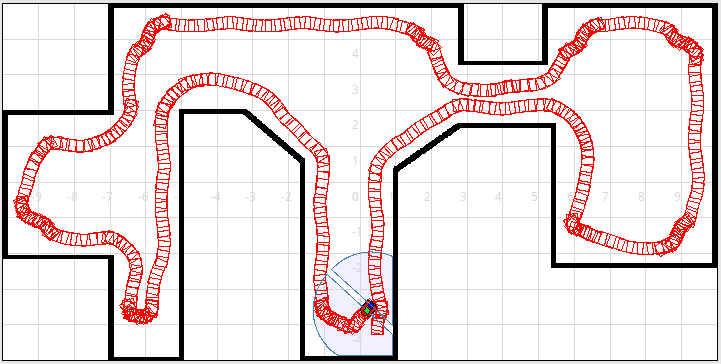}}\\
\end{tabular}
\caption{\label{Fig:environments1}Path of the robot along the simulated environments (I).}
\end{figure*}

\begin{figure*}[tb!]
\centering
\begin{tabular}{>{\centering\arraybackslash}m{0.28\textwidth}>{\centering\arraybackslash}m{0.28\textwidth}>{\centering\arraybackslash}m{0.28\textwidth}}
\subfigure[\emph{rooms}\label{Fig:rooms}]{\includegraphics[scale=0.22]{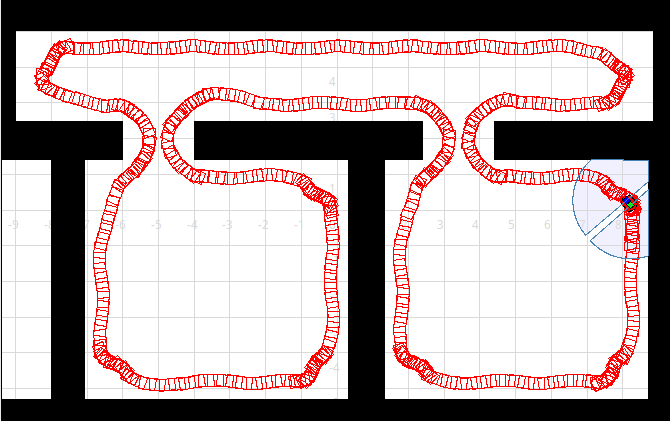}} &
\subfigure[\emph{flower}\label{Fig:flower}]{\includegraphics[scale=0.22]{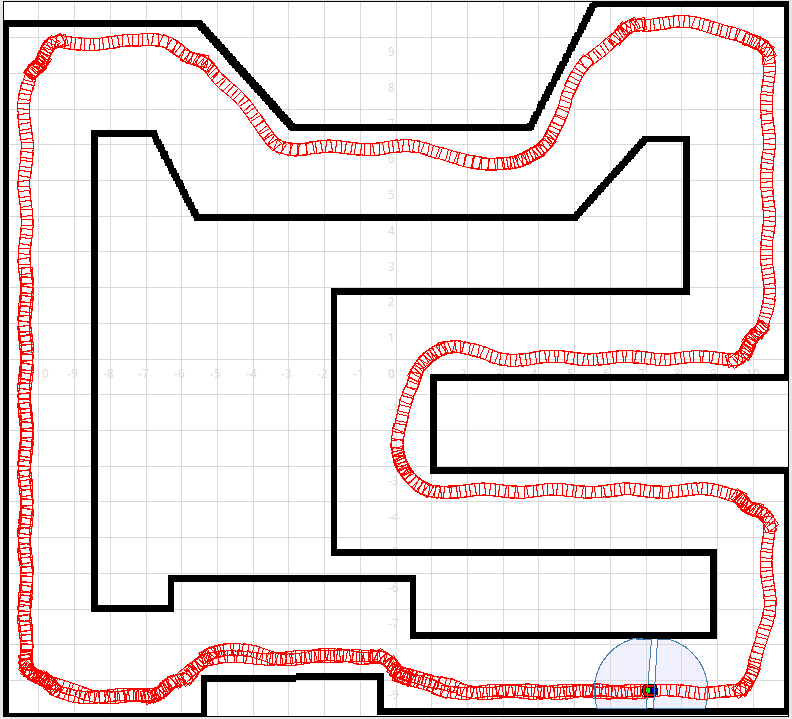}} & 
\subfigure[\emph{office}\label{Fig:office}]{\includegraphics[scale=0.22]{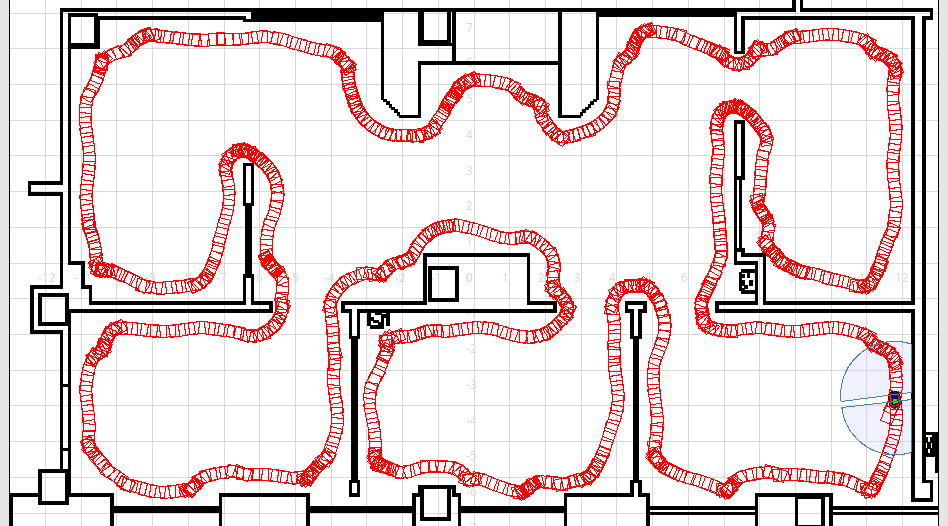}}\\
\multicolumn{2}{c}{\addtolength{\subfigcapskip}{0.1cm} \subfigure[\emph{autolab}\label{Fig:autolab}]{\includegraphics[scale=0.22]{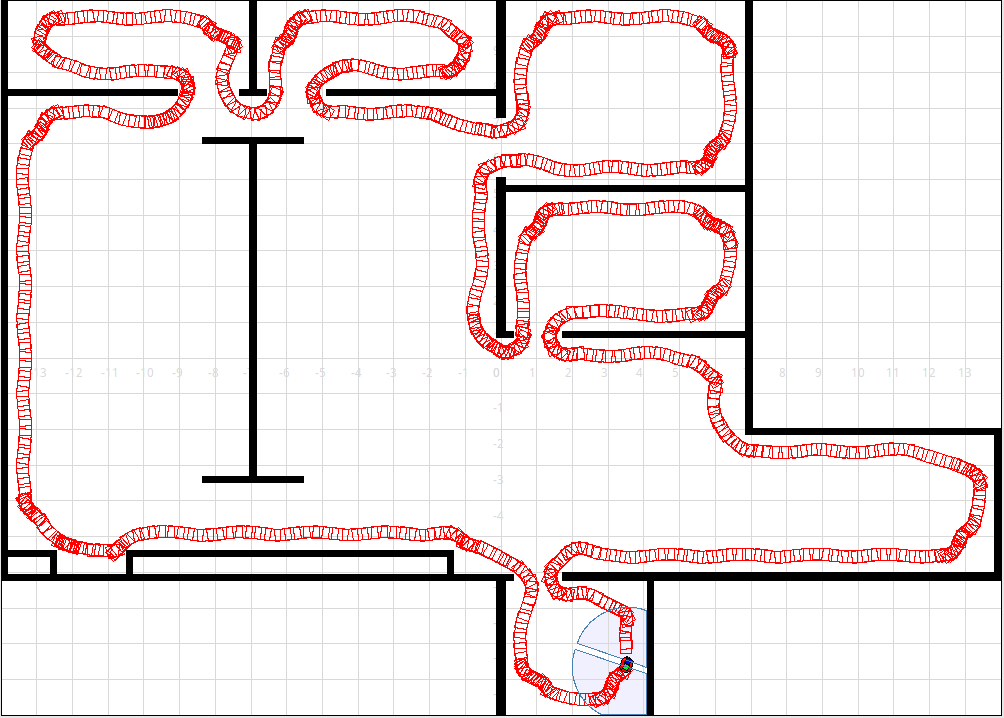}}} &
\multicolumn{1}{c}{\addtolength{\subfigcapskip}{0.1cm} \subfigure[\emph{maze}\label{Fig:maze}]{\includegraphics[scale=0.22]{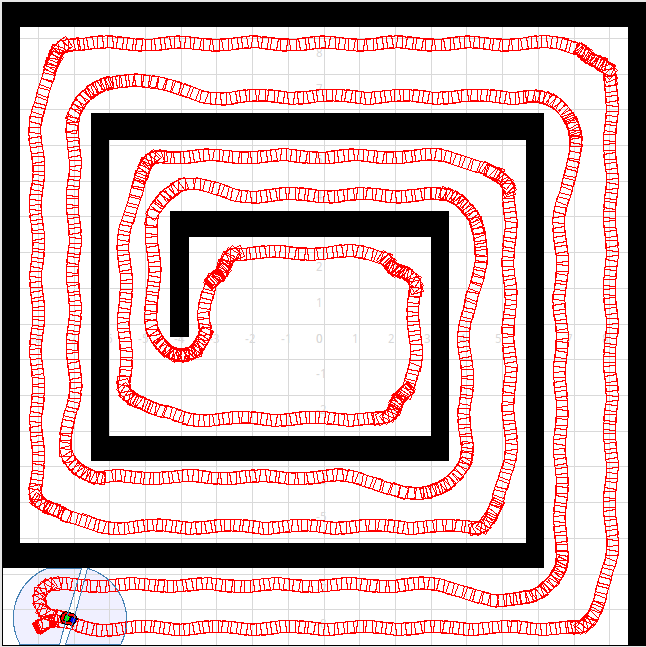}}}\\
\multicolumn{3}{c}{\addtolength{\subfigcapskip}{0.2cm} \subfigure[\emph{hospital}\label{Fig:hospital}]{\includegraphics[scale=0.22]{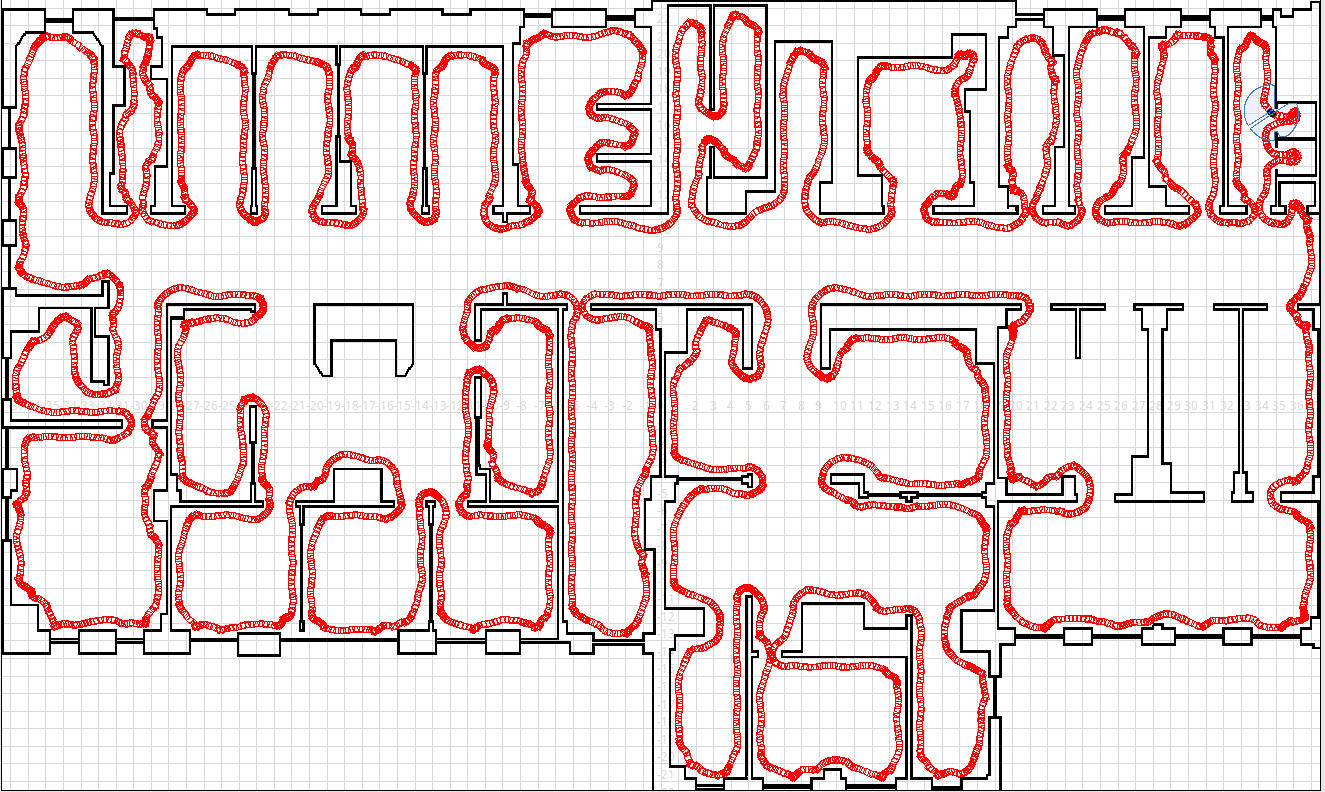}}}\\
\end{tabular}
\caption{\label{Fig:environments2}Path of the robot along the simulated environments (II).}
\end{figure*}

\begin{figure*}[tb!]
\centering
\subfigure[\emph{real environment 1}\label{Fig:real1}]{\includegraphics[scale=0.3]{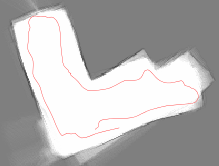}}\hspace*{1cm}
\subfigure[\emph{real environment 2}\label{Fig:real2}]{\includegraphics[scale=0.3]{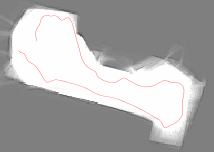}}\\
\caption{\label{Fig:realenvironments}Path of the robot along the real environments.}
\end{figure*}

The simulated environments are shown in Figs. \ref{Fig:environments1} and \ref{Fig:environments2}. The trace of the robot is represented by marks, and the higher the concentration of marks, the lower the velocity of the robot. Furthermore, Fig. \ref{Fig:realenvironments} shows the real environments. Each of them represents an occupancy grid map of the environment, together with the trajectory of the robot.

\subsection{Algorithms and parameters \label{Sec:parameters}}
The following values were used for the parameters of the evolutionary algorithm: $\textnormalit{ME} = 0.02$, $\textnormalit{DOF}_{min} = 0.001$, $\alpha_f = 0.99$, $P_{cross} = 0.8$, $\textnormalit{pop}_{max} = 70$, $\textnormalit{it}_{min} = 50$, $\textnormalit{it}_{check} = 10$, $\textnormalit{it}_{max} = 100$, $\sigma_{bd}=0.01$, $\sigma_{v} = 0.1$ and $P_{min} = 0.17$. $P_{min}$ is a parameter that has a high influence in the performance of the system. A single value of $P_{min}$ was used in testing, obtained from Eqs. \ref{Eq:probEx} and \ref{Eq:error}  for the case the error for each consequent is one label (Eq. \ref{Eq:error}). The granularities and the universe of discourse of each output of a rule are shown in table \ref{Tab:universes}. For the rule subset selection algorithm, the parameters have values of $radius_{nbhood} = 1$ and $maxRestarts = 2$.

The fuzzy inference system used for the learned fuzzy rule sets uses the minimum t-norm for both the implication and conjunction operators, and the weighted average method as defuzzification operator.

\begin{table}[tb!]
\centering
\small
\caption{\label{Tab:universes}Universe of discourse and granularities}
\begin{tabular}{|c|c|c|c|}
\hline
Variable & Min & Max & Granularities\\
\hline
\hline
Distance & $0$ & $1.5$ & All\\
Beam & $0$ & $721$ & All\\
Quantifier & $10$ & $100$ & $-$\\
Velocity & $0$ & $0.5$ & All\\
Lineal velocity & $0$ & $0.5$ & $\{9\}$\\
Angular velocity & $-\pi/4$ & $\pi/4$ & $\{19\}$\\
\hline
\end{tabular}
\end{table}

The IQFRL approach was compared with three different algorithms:
\begin{itemize}
\item Methodology to Obtain Genetic fuzzy rule-based systems Under the iterative Learning approach (MOGUL): a three-stage genetic algorithm \cite{cordón1997_ar}:
\begin{enumerate}
    \item An evolutionary process for learning fuzzy rules, with two components: a fuzzy-rule generating method based on IRL, and an iterative covering method.
    \item A genetic simplification process for selecting rules.
    \item A genetic tuning process, that tunes the membership functions for each fuzzy rule or for the complete rule base.
\end{enumerate}
The soft-constrained MOGUL was used, as it has better performance in very hard problems \cite{cordón2001_fss}\footnote{The implementation in \textit{Keel} \cite{alcalá2009_keel}, an open source (GPLv3) Java software tool to assess evolutionary algorithms for Data Mining problems, was used.}.

\item Multilayer Perceptron Neural Network (MPNN): a single-hidden-layer neural network trained with the BFGS method \cite{setiono1995use} with the following parameters: $abstol = 0.01$, $reltol = 0.0001$ and $maxit = 500$. The number of neurons in the hidden layer varies from $n$ to $2\cdot n$, being $n$ the number of inputs\footnote{The package \textit{nnet} \cite{venables2002} of the statistical software R was used.}.

\item $\nu$-Support Vector Regression ($\nu$-SVR)\footnote{The package \textit{kernlab} \cite{karatzoglou2004kernlab} of the statistical software R was used.}: a $\nu$-SVM \cite{schölkopf2000_nc} version for regression with a Gaussian RBF kernel. The parameter sigma is estimated based upon the 0.1 and 0.9 quantile of $||x\ -\ x'||^2$.
\end{itemize}

As mentioned before, in the IQFRL proposal the preprocessing of raw sensor data is embedded in the learning algorithm. Since the algorithms for the comparison need to preprocess the data before the learning phase, three different approaches were used for the transformation of the sensor data:
\begin{itemize}
 \item \textit{Min}: the beams of the laser range finder are grouped in $n$ equal sized sectors. For each sector, the minimum distance value is selected as input.
 \item \textit{Sample}: $n$ equidistant beams are selected as the input data.
 \item PCA: Principal Component Analysis computes the most meaningful basis to re-express the data. It is a simple, non-parametric method for extracting relevant information. The variances associated with the principal components can be examined in order to select only those that cover a percentage of the total variance. 
\end{itemize}
Different parameters have been used for the preprocessing approaches. For \textit{Min} and \textit{Sample} methods, the number of obtained inputs ($n$) was changed. For \textit{PCA}, the percentage of variance ($\sigma_{PCA}$) indicates the principal components selected as input data. Table \ref{Tab:sampling} shows the parameters used for the preprocessing methods. Moreover, table \ref{Tab:PCA} shows the number of inputs obtained with PCA for the three datasets with each configuration.

\begin{table}[tb!]
\centering
\small
\caption{\label{Tab:sampling}Different configurations for the preprocessing methods.}
\begin{tabular}{|c|c|c|c|c|c|}
\hline
Preprocessing & Configuration \\
\hline
\hline
Min (n) & $\{4, 8, 16, 32, 64\}$ \\
Sample (n) & $\{4, 8, 16, 32, 64\}$ \\
PCA ($\sigma_{PCA}$) & $\{0.90, 0.95, 0.975, 0.99, 0.999\}$ \\
\hline
\end{tabular}
\end{table}

\begin{table}[tb!]
\centering
\small
\caption{\label{Tab:PCA}Number of inputs obtained with PCA.}
\begin{tabular}{|c|c|c|c|}
\hline
$\sigma_{PCA}$ & Straight & Convex & Concave \\
\hline
\hline
$0.90$ & $35$ & $15$ & $27$\\
$0.95$ & $51$ & $24$ & $40$\\
$0.975$ & $66$ & $35$ & $53$\\
$0.99$ & $85$ & $57$ & $68$\\
$0.999$ & $127$ & $99$ & $109$\\
\hline
\end{tabular}
\end{table}

\subsection{\label{sec:comparison}Comparison and statistical significance}
Table \ref{Tab:error} shows the training and test errors over a 5-fold cross-validation. For each algorithm and dataset the mean and standard deviation of the error (Eq. \ref{Eq:error}) were calculated.

\begin{table}[tb!]
\centering
\small
\caption{\label{Tab:error}Training and test errors}
\begin{tabular}{|c|c|c|c|c|}
\hline
Alg. & Preproc. & Dataset & Training & Test\\
\hline
\hline
\multirow{3}{*}{IQFRL} & $-$ & Straight & $0.11 \pm 0.03$ & $0.14 \pm 0.03$\\
& & Convex & $0.10 \pm 0.01$ & $0.12 \pm 0.02$\\
& & Concave & $0.04 \pm 0.01$ & $0.05 \pm 0.01$\\
\hline
\multirow{6}{*}{MOGUL} & \multirow{3}{*}{min 16}  & other &  $0.01 \pm 0.00$ &  $0.10 \pm 0.01$\\
& & convex &  $0.01 \pm 0.00$ &  $0.05 \pm 0.01$\\
& & concave &  $0.00 \pm 0.00$ &  $0.05 \pm 0.01$\\
\cline{2-5}
 & \multirow{3}{*}{sample 16}  & other &  $0.01 \pm 0.00$ &  $0.09 \pm 0.02$\\
& & convex &  $0.02 \pm 0.00$ &  $0.05 \pm 0.01$\\
& & concave &  $0.00 \pm 0.00$ &  $0.04 \pm 0.01$\\
\hline
\multirow{6}{*}{MPNN} & \multirow{3}{*}{min 8} & other & $ 0.01 \pm 0.00 $ & $ 0.06 \pm 0.10$\\
& & convex & $ 0.02 \pm 0.01 $ & $ 0.03 \pm 0.05$\\
& & concave & $ 0.00 \pm 0.00 $ & $ 0.04 \pm 0.06$\\
\cline{2-5}
 & \multirow{3}{*}{sample 8} & other & $ 0.02 \pm 0.00 $ & $ 0.18 \pm 0.25$\\
& & convex & $ 0.03 \pm 0.01 $ & $ 0.02 \pm 0.02$\\
& & concave & $ 0.01 \pm 0.00 $ & $ 0.17 \pm 0.34$\\
\hline
\multirow{6}{*}{$\nu$-SVR} & \multirow{3}{*}{min 16} & other & $ 0.01 \pm 0.00 $ & $ 0.02 \pm 0.02$\\
& & convex & $ 0.03 \pm 0.01 $ & $ 0.02 \pm 0.01$\\
& & concave & $ 0.01 \pm 0.00 $ & $ 0.00 \pm 0.00$\\
\cline{2-5}
 & \multirow{3}{*}{sample 16} & other & $ 0.02 \pm 0.01 $ & $ 0.02 \pm 0.02$\\
& & convex & $ 0.04 \pm 0.02 $ & $ 0.02 \pm 0.01$\\
& & concave & $ 0.01 \pm 0.00 $ & $ 0.01 \pm 0.00$\\
\cline{2-5}

\hline
\end{tabular}
\end{table}

For each preprocessing technique, a 5-fold cross-validation was performed for each combination of the parameters of the algorithms. For example, for the \textit{Min} preprocessing with 16 equal size sectors, a 5-fold cross-validation was run for each number of neurons between 17 and 34 for the MPNN approach. Only the configuration of the algorithm with lowest test error for each configuration of the preprocessing methods was used for comparison purposes. Moreover, only those configurations of preprocessing techniques with the best results are shown in the tables of this section. Results for \textit{PCA} preprocessing have not been included, as the learning algorithms were not able to obtain adequate controllers.

Although, the MSE (Mean Squared Error) is the usual measure of the performance of the algorithms, this is not a sufficient criterion in mobile robotics. A good controller must be robust and able to provide a good and smooth output in any situation. The only way to validate the controller is to test it on environments (simulated and real) with different difficulties and assessing on these tests a number of quality parameters such as mean distance to the wall, mean velocity along the paths, \dots 

Table \ref{Tab:results} contains the results of the execution of each of the algorithms for the different simulated environments (Figs. \ref{Fig:environments1} and \ref{Fig:environments2}). Furthermore, table \ref{Tab:average_results} shows the average results for the following five different indicators: the distance to the wall at its right (Dist.), the linear velocity (Vel.), the change in the linear velocity between two consecutive cycles (Vel.ch.) ---which reflects the smoothness in the control---, the time per lap, and the number of blockades of the robot along the path and cannot recover.
 
The robot is blocked if it hits a wall or if it does not move for 5 s. In this situation the robot is placed parallel to the wall at a distance of 0.5 m. The average values of the five indicators are calculated for each lap that the robot performs in the environment. Results presented in the table are the average and standard deviation values over five laps of the average values of the indicators over one lap. The dash symbol in the results table indicates that the controller could not complete the path. This usually occurs when the number of blockades per meter is high (greater than 5 blockades in a short period of time) or when the robot completely deviates from the path.

Moreover, in order to evaluate the performance of a controller with a numerical value a general quality measure was defined. It is based on the error measure defined in \cite{Mucientes10_eswa}, but including the number of blockades:
\begin{equation}
\scriptsize
 quality = \frac{1}{1+(1+\# Blockades) \cdot \left(0.9 \cdot |Dist - d_{wall}| + 0.1 \cdot |Vel - v_{max}|\right)}
\end{equation}
where $d_{wall}$ is the reference distance to the wall (50 cm) and $v_{max}$ is the maximum value of the velocity (50 cm/s). The higher the quality, the better the controller. This measure takes the number of blockades into account in a linear form for comparison purposes. However, it should be noted that controllers with just a single blockade are not reliable and should not be implemented on a real robot.  

\begin{table}[b!]
\centering
\small
\caption{\label{Tab:quality}Non-parametric test for $quality$ of table \ref{Tab:results}.}
\begin{tabular}{|c|c|c|c|}
\hline
Alg. & Preprocessing & Ranking & Holm \textit{p}-value\\
\hline
\hline
IQFRL & $-$ & $1.53$ & $-$\\
\hline
\multirow{2}{*}{MOGUL} & min 16 & $4.9$ & $0.012$\\
\cline{2-4}
 & sample 16 & $3.47$ & $0.025$\\
\hline
\multirow{2}{*}{MLPNN} & min 8 & $5.57$ & $0.010$\\
\cline{2-4}
 & sample 8 & $6$ & $0.008$\\
\hline
\multirow{2}{*}{$\nu$-SVR} & min 16 & $3.83$ & $0.017$\\
\cline{2-4}
 & sample 16 & $2.7$ & $0.05$\\
\hline
\multicolumn{4}{|c|}{Friedman \textit{p}-value = $0.00$}\\
\multicolumn{4}{|c|}{Holm's rejects hypothesis with \textit{p}-value $<= 0.05$}\\
\hline
\end{tabular}
\end{table}

\begin{table*}[tb!]
\ssmall
\renewcommand{\arraystretch}{0.88}
\centering
\caption{\label{Tab:results}Average results ($x \pm \sigma$) for each simulated environment}
\begin{tabular}{|c|c|c|c|c|c|c|c|Hc|}

\hline
Alg. & Prepr. & Env. & Dist.(cm) & Vel.(cm/s) & Vel.ch.(cm/s) & Time(s) & \# Blockades & $error$ & $quality$\\
\hline
\hline

\multirow{15}{*}{IQFRL} & \multirow{15}{*}{$-$} & home & $55.70 \pm 0.25$ & $27.00 \pm 0.66$ & $5.59 \pm 0.14$ & $164.63 \pm 5.26$ & $0.00 \pm 0.00$ & $7.43$ & $0.12$ \\
 &  & gfs\_b & $55.98 \pm 1.70$ & $22.37 \pm 0.92$ & $7.01 \pm 0.74$ & $163.90 \pm 9.72$ & $0.00 \pm 0.00$ & $8.14$ & $0.11$ \\
 &  & dec & $57.33 \pm 1.02$ & $32.47 \pm 0.67$ & $5.83 \pm 0.18$ & $168.63 \pm 1.76$ & $0.00 \pm 0.00$ & $8.35$ & $0.11$ \\
 &  & domus & $54.84 \pm 0.53$ & $29.97 \pm 0.19$ & $5.97 \pm 0.49$ & $198.80 \pm 1.39$ & $0.00 \pm 0.00$ & $6.36$ & $0.14$ \\
 &  & citius & $54.80 \pm 0.87$ & $26.11 \pm 0.78$ & $6.28 \pm 0.64$ & $249.50 \pm 8.25$ & $0.00 \pm 0.00$ & $6.71$ & $0.13$ \\
 &  & raid\_a & $59.56 \pm 0.72$ & $25.35 \pm 0.14$ & $6.87 \pm 0.55$ & $262.00 \pm 6.15$ & $0.00 \pm 0.00$ & $11.07$ & $0.08$ \\
 &  & wsc8a & $56.96 \pm 1.00$ & $27.45 \pm 0.84$ & $7.70 \pm 0.33$ & $233.10 \pm 5.28$ & $0.00 \pm 0.00$ & $8.52$ & $0.11$ \\
 &  & home\_b & $58.41 \pm 0.72$ & $25.53 \pm 0.46$ & $7.06 \pm 0.36$ & $300.07 \pm 8.15$ & $0.00 \pm 0.00$ & $10.02$ & $0.09$ \\
 &  & raid\_b & $58.22 \pm 0.44$ & $28.57 \pm 0.40$ & $6.60 \pm 0.47$ & $242.23 \pm 3.82$ & $0.00 \pm 0.00$ & $9.54$ & $0.09$ \\
 &  & rooms & $57.38 \pm 0.34$ & $30.97 \pm 0.34$ & $6.38 \pm 0.43$ & $261.93 \pm 4.60$ & $0.00 \pm 0.00$ & $8.54$ & $0.10$ \\
 &  & flower & $53.46 \pm 0.25$ & $33.85 \pm 0.40$ & $4.13 \pm 0.34$ & $290.77 \pm 4.13$ & $0.00 \pm 0.00$ & $4.73$ & $0.17$ \\
 &  & office & $51.37 \pm 0.57$ & $24.20 \pm 0.18$ & $6.65 \pm 0.25$ & $578.27 \pm 2.92$ & $0.00 \pm 0.00$ & $3.81$ & $0.21$ \\
 &  & autolab & $52.91 \pm 0.20$ & $28.75 \pm 0.31$ & $5.57 \pm 0.48$ & $499.33 \pm 9.74$ & $0.00 \pm 0.00$ & $4.74$ & $0.17$ \\
 &  & maze & $52.43 \pm 0.22$ & $35.88 \pm 0.40$ & $3.64 \pm 0.28$ & $567.73 \pm 5.29$ & $0.00 \pm 0.00$ & $3.60$ & $0.22$ \\
 &  & hospital & $51.09 \pm 0.19$ & $26.68 \pm 0.10$ & $6.18 \pm 0.35$ & $3608.07 \pm 21.72$ & $0.00 \pm 0.00$ & $3.31$ & $0.23$ \\
\hline
\multirow{30}{*}{MOGUL} & \multirow{15}{*}{min 16}  & home & $55.12 \pm 0.69$ & $30.43 \pm 1.30$ & $5.40 \pm 0.45$ & $181.10 \pm 12.70$ & $7.33 \pm 2.05$ & $6.56$ & $0.05$ \\
 & & gfs\_b & $54.75 \pm 0.96$ & $24.44 \pm 1.02$ & $6.81 \pm 0.39$ & $208.87 \pm 12.86$ & $14.00 \pm 2.16$ & $6.83$ & $0.04$ \\
 & & dec & $55.46 \pm 1.14$ & $36.13 \pm 0.43$ & $5.17 \pm 0.15$ & $190.50 \pm 6.29$ & $8.67 \pm 1.25$ & $6.30$ & $0.07$ \\
 & & domus & $55.75 \pm 0.64$ & $31.44 \pm 1.80$ & $5.47 \pm 0.33$ & $224.60 \pm 10.25$ & $8.33 \pm 0.47$ & $7.03$ & $0.09$ \\
 & & citius & $53.47 \pm 1.27$ & $29.48 \pm 0.13$ & $5.60 \pm 0.53$ & $302.57 \pm 13.58$ & $18.33 \pm 2.62$ & $5.17$ & $0.05$ \\
 & & raid\_a & $57.53 \pm 0.32$ & $26.53 \pm 0.28$ & $6.41 \pm 0.09$ & $363.87 \pm 10.21$ & $27.33 \pm 3.40$ & $9.13$ & $0.02$ \\
 & & wsc8a & $54.80 \pm 0.35$ & $27.57 \pm 0.63$ & $6.26 \pm 0.59$ & $346.90 \pm 29.40$ & $26.67 \pm 5.44$ & $6.57$ & $0.02$ \\
 & & home\_b & $56.75 \pm 0.49$ & $27.49 \pm 0.62$ & $6.58 \pm 0.37$ & $379.57 \pm 2.05$ & $22.00 \pm 1.41$ & $8.33$ & $0.05$ \\
 & & raid\_b & $57.48 \pm 0.70$ & $32.38 \pm 0.17$ & $5.84 \pm 0.38$ & $280.17 \pm 14.29$ & $14.67 \pm 3.40$ & $8.50$ & $0.03$ \\
 & & rooms & $54.79 \pm 0.38$ & $30.57 \pm 1.04$ & $5.14 \pm 0.37$ & $350.33 \pm 28.04$ & $20.00 \pm 5.89$ & $6.25$ & $0.02$ \\
 & & flower & $53.33 \pm 0.39$ & $38.05 \pm 1.00$ & $3.70 \pm 0.71$ & $310.27 \pm 13.41$ & $11.67 \pm 4.03$ & $4.19$ & $0.05$ \\
 & & office & $51.48 \pm 0.29$ & $25.09 \pm 0.49$ & $6.77 \pm 0.20$ & $762.20 \pm 5.28$ & $49.67 \pm 1.25$ & $3.82$ & $0.10$ \\
 & & autolab & $51.95 \pm 0.71$ & $30.54 \pm 1.24$ & $5.23 \pm 0.27$ & $612.00 \pm 23.46$ & $31.00 \pm 2.45$ & $3.70$ & $0.07$ \\
 & & maze & $52.25 \pm 0.55$ & $37.55 \pm 1.53$ & $2.87 \pm 0.25$ & $690.93 \pm 53.92$ & $32.00 \pm 6.38$ & $3.27$ & $0.04$ \\
 & & hospital & $51.33 \pm 0.06$ & $26.86 \pm 0.08$ & $5.89 \pm 0.29$ & $4908.07 \pm 56.12$ & $313.33 \pm 10.34$ & $3.51$ & $0.02$ \\
\cline{2-10}
 & \multirow{15}{*}{sample 16} & home & $56.76 \pm 0.20$ & $29.57 \pm 0.35$ & $4.73 \pm 0.15$ & $161.97 \pm 0.69$ & $1.67 \pm 0.47$ & $8.13$ & $0.08$ \\
 & & gfs\_b & $56.16 \pm 1.62$ & $23.66 \pm 0.88$ & $8.16 \pm 0.51$ & $160.80 \pm 9.65$ & $1.67 \pm 1.25$ & $8.18$ & $0.05$ \\
 & & dec & $57.69 \pm 0.66$ & $37.95 \pm 0.96$ & $6.03 \pm 0.25$ & $148.87 \pm 3.94$ & $0.67 \pm 0.47$ & $8.13$ & $0.08$ \\
 & & domus & $56.04 \pm 0.20$ & $36.61 \pm 1.14$ & $6.35 \pm 0.57$ & $165.63 \pm 8.93$ & $1.00 \pm 0.82$ & $6.77$ & $0.08$ \\
 & & citius & $51.38 \pm 1.72$ & $27.40 \pm 0.21$ & $6.13 \pm 0.52$ & $241.53 \pm 3.38$ & $1.00 \pm 0.82$ & $3.50$ & $0.14$ \\
 & & raid\_a & $57.44 \pm 0.51$ & $26.18 \pm 1.39$ & $6.61 \pm 0.44$ & $275.63 \pm 18.82$ & $3.67 \pm 2.05$ & $9.08$ & $0.03$ \\
 & & wsc8a & $54.67 \pm 0.24$ & $30.18 \pm 0.65$ & $9.03 \pm 0.42$ & $220.87 \pm 2.26$ & $1.67 \pm 0.94$ & $6.18$ & $0.08$ \\
 & & home\_b & $57.17 \pm 0.36$ & $26.80 \pm 0.91$ & $6.73 \pm 0.72$ & $303.77 \pm 9.52$ & $3.00 \pm 0.82$ & $8.78$ & $0.06$ \\
 & & raid\_b & $60.38 \pm 0.38$ & $34.73 \pm 1.26$ & $6.15 \pm 0.41$ & $206.20 \pm 10.12$ & $0.33 \pm 0.47$ & $10.87$ & $0.06$ \\
 & & rooms & $56.05 \pm 0.09$ & $32.11 \pm 1.49$ & $6.39 \pm 0.64$ & $254.50 \pm 18.03$ & $0.67 \pm 0.94$ & $7.23$ & $0.07$ \\
 & & flower & $55.24 \pm 0.58$ & $41.58 \pm 0.32$ & $3.92 \pm 0.27$ & $244.67 \pm 6.56$ & $2.00 \pm 1.41$ & $5.56$ & $0.07$ \\
 & & office & $50.33 \pm 0.10$ & $22.76 \pm 0.56$ & $6.22 \pm 0.46$ & $655.40 \pm 21.32$ & $11.33 \pm 2.36$ & $3.02$ & $0.09$ \\
 & & autolab & $50.57 \pm 0.27$ & $29.62 \pm 0.69$ & $5.29 \pm 0.32$ & $498.67 \pm 8.12$ & $2.33 \pm 1.25$ & $2.55$ & $0.15$ \\
 & & maze & $55.05 \pm 0.34$ & $40.22 \pm 0.78$ & $3.25 \pm 0.18$ & $512.33 \pm 8.86$ & $0.67 \pm 0.47$ & $5.52$ & $0.11$ \\
 & & hospital & $51.54 \pm 0.25$ & $25.97 \pm 0.14$ & $6.11 \pm 0.34$ & $3964.70 \pm 15.81$ & $64.67 \pm 8.18$ & $3.79$ & $0.03$ \\
\hline
\multirow{30}{*}{MPNN} & \multirow{15}{*}{min 8}  & home & $58.34 \pm 1.07$ & $29.34 \pm 2.66$ & $4.12 \pm 0.11$ & $122.73 \pm 6.36$ & $6.33 \pm 0.47$ & $9.57$ & $0.07$ \\
 & & gfs\_b & $58.56 \pm 1.41$ & $28.14 \pm 0.71$ & $7.66 \pm 0.28$ & $149.20 \pm 11.03$ & $4.33 \pm 1.70$ & $9.89$ & $0.04$ \\
 & & dec & $56.10 \pm 0.32$ & $35.13 \pm 0.38$ & $4.28 \pm 0.19$ & $173.37 \pm 5.58$ & $3.33 \pm 0.94$ & $6.97$ & $0.07$ \\
 & & domus & $-$ & $-$ & $-$ & $-$ & $-$ & $-$ & $0.00$ \\
 & & citius & $55.73 \pm 0.90$ & $27.22 \pm 0.68$ & $4.97 \pm 0.28$ & $324.37 \pm 31.51$ & $15.00 \pm 3.56$ & $7.44$ & $0.03$ \\
 & & raid\_a & $57.90 \pm 0.61$ & $23.05 \pm 1.01$ & $6.85 \pm 0.26$ & $403.60 \pm 13.17$ & $27.33 \pm 1.70$ & $9.80$ & $0.04$ \\
 & & wsc8a & $56.24 \pm 0.81$ & $30.96 \pm 1.03$ & $7.97 \pm 0.11$ & $238.77 \pm 5.45$ & $7.33 \pm 0.47$ & $7.52$ & $0.08$ \\
 & & home\_b & $58.02 \pm 0.54$ & $24.19 \pm 1.81$ & $7.13 \pm 0.45$ & $922.87 \pm 568.90$ & $66.33 \pm 41.02$ & $9.80$ & $0.00$ \\
 & & raid\_b & $59.99 \pm 1.57$ & $27.98 \pm 2.64$ & $5.07 \pm 0.63$ & $1137.37 \pm 842.79$ & $38.33 \pm 20.53$ & $11.19$ & $0.00$ \\
 & & rooms & $-$ & $-$ & $-$ & $-$ & $-$ & $-$ & $0.00$ \\
 & & flower & $-$ & $-$ & $-$ & $-$ & $-$ & $-$ & $0.00$ \\
 & & office & $55.84 \pm 0.48$ & $28.48 \pm 0.19$ & $8.18 \pm 0.32$ & $626.33 \pm 5.59$ & $32.00 \pm 1.41$ & $7.40$ & $0.05$ \\
 & & autolab & $-$ & $-$ & $-$ & $-$ & $-$ & $-$ & $0.00$ \\
 & & maze & $52.75 \pm 0.32$ & $42.53 \pm 1.09$ & $2.81 \pm 0.38$ & $621.07 \pm 45.13$ & $28.00 \pm 3.56$ & $3.22$ & $0.06$ \\
 & & hospital & $55.94 \pm 0.02$ & $28.45 \pm 0.33$ & $7.47 \pm 0.17$ & $3730.00 \pm 166.69$ & $205.00 \pm 13.93$ & $7.50$ & $0.01$ \\
\cline{2-10}
 & \multirow{15}{*}{sample 8}  & home & $-$ & $-$ & $-$ & $-$ & $-$ & $-$ & $0.00$ \\
 & & gfs\_b & $62.11 \pm 0.47$ & $21.96 \pm 0.35$ & $7.22 \pm 0.14$ & $172.00 \pm 0.78$ & $1.00 \pm 0.00$ & $13.71$ & $0.07$ \\
 & & dec & $64.67 \pm 2.80$ & $30.23 \pm 4.21$ & $6.03 \pm 0.49$ & $285.70 \pm 118.55$ & $1.00 \pm 0.82$ & $15.18$ & $0.04$ \\
 & & domus & $-$ & $-$ & $-$ & $-$ & $-$ & $-$ & $0.00$ \\
 & & citius & $68.36 \pm 5.48$ & $20.98 \pm 2.23$ & $6.69 \pm 0.18$ & $603.33 \pm 243.12$ & $16.33 \pm 11.81$ & $19.43$ & $0.00$ \\
 & & raid\_a & $74.58 \pm 8.62$ & $19.36 \pm 3.61$ & $6.41 \pm 1.03$ & $450.60 \pm 259.59$ & $5.00 \pm 3.56$ & $25.18$ & $0.01$ \\
 & & wsc8a & $61.04 \pm 0.62$ & $23.70 \pm 0.41$ & $7.90 \pm 0.52$ & $279.37 \pm 6.70$ & $1.00 \pm 0.82$ & $12.57$ & $0.04$ \\
 & & home\_b & $85.20 \pm 8.36$ & $16.95 \pm 3.09$ & $6.40 \pm 0.62$ & $2477.03 \pm 1471.27$ & $26.33 \pm 15.15$ & $34.98$ & $0.00$ \\
 & & raid\_b & $70.10 \pm 4.50$ & $21.41 \pm 1.56$ & $7.18 \pm 0.48$ & $1780.97 \pm 1445.69$ & $49.00 \pm 43.69$ & $20.95$ & $0.00$ \\
 & & rooms & $60.75 \pm 0.47$ & $34.39 \pm 0.65$ & $6.81 \pm 0.22$ & $237.53 \pm 5.88$ & $0.00 \pm 0.00$ & $11.24$ & $0.08$ \\
 & & flower & $61.77 \pm 2.11$ & $28.82 \pm 0.35$ & $7.54 \pm 0.21$ & $912.67 \pm 231.62$ & $51.33 \pm 13.02$ & $12.71$ & $0.01$ \\
 & & office & $57.22 \pm 1.31$ & $21.01 \pm 0.31$ & $5.58 \pm 0.15$ & $783.03 \pm 7.34$ & $28.33 \pm 0.47$ & $9.40$ & $0.07$ \\
 & & autolab & $-$ & $-$ & $-$ & $-$ & $-$ & $-$ & $0.00$ \\
 & & maze & $-$ & $-$ & $-$ & $-$ & $-$ & $-$ & $0.00$ \\
 & & hospital & $74.51 \pm 11.21$ & $21.74 \pm 3.14$ & $5.33 \pm 1.18$ & $555.43 \pm 721.71$ & $16.33 \pm 23.10$ & $24.89$ & $0.00$ \\
\hline
\multirow{30}{*}{$\nu$-SVR} & \multirow{15}{*}{min 16} & home & $-$ & $-$ & $-$ & $-$ & $-$ & $-$ & $0.00$ \\
 & & gfs\_b & $57.82 \pm 0.58$ & $26.35 \pm 0.81$ & $8.67 \pm 0.52$ & $140.20 \pm 3.94$ & $0.00 \pm 0.00$ & $9.40$ & $0.10$ \\
 & & dec & $59.14 \pm 0.05$ & $39.01 \pm 0.98$ & $6.59 \pm 0.48$ & $143.03 \pm 3.39$ & $0.00 \pm 0.00$ & $9.32$ & $0.10$ \\
 & & domus & $-$ & $-$ & $-$ & $-$ & $-$ & $-$ & $0.00$ \\
 & & citius & $55.14 \pm 0.78$ & $25.65 \pm 0.29$ & $5.90 \pm 0.28$ & $258.87 \pm 2.23$ & $0.00 \pm 0.00$ & $7.06$ & $0.12$ \\
 & & raid\_a & $58.52 \pm 0.29$ & $30.31 \pm 0.67$ & $9.36 \pm 0.28$ & $208.97 \pm 4.84$ & $0.00 \pm 0.00$ & $9.64$ & $0.09$ \\
 & & wsc8a & $58.33 \pm 0.24$ & $30.09 \pm 0.27$ & $10.70 \pm 0.38$ & $218.33 \pm 1.56$ & $0.00 \pm 0.00$ & $9.48$ & $0.10$ \\
 & & home\_b & $61.26 \pm 0.46$ & $27.87 \pm 0.50$ & $8.05 \pm 0.10$ & $289.40 \pm 2.94$ & $1.00 \pm 0.00$ & $12.34$ & $0.07$ \\
 & & raid\_b & $-$ & $-$ & $-$ & $-$ & $-$ & $-$ & $0.00$ \\
 & & rooms & $-$ & $-$ & $-$ & $-$ & $-$ & $-$ & $0.00$ \\
 & & flower & $58.66 \pm 0.11$ & $38.42 \pm 0.64$ & $4.76 \pm 0.22$ & $257.10 \pm 4.64$ & $0.00 \pm 0.00$ & $8.95$ & $0.10$ \\
 & & office & $51.37 \pm 0.47$ & $23.92 \pm 0.40$ & $7.67 \pm 0.10$ & $582.23 \pm 5.47$ & $0.00 \pm 0.00$ & $3.84$ & $0.21$ \\
 & & autolab & $54.18 \pm 0.25$ & $28.04 \pm 0.22$ & $6.38 \pm 0.19$ & $522.07 \pm 3.83$ & $0.67 \pm 0.47$ & $5.95$ & $0.10$ \\
 & & maze & $61.07 \pm 0.70$ & $32.60 \pm 1.02$ & $3.06 \pm 0.17$ & $675.67 \pm 63.68$ & $1.00 \pm 0.82$ & $11.70$ & $0.04$ \\
 & & hospital & $53.42 \pm 0.36$ & $25.42 \pm 0.08$ & $6.58 \pm 0.21$ & $3833.90 \pm 13.48$ & $6.67 \pm 1.70$ & $5.53$ & $0.06$ \\
\cline{2-10}
 & \multirow{15}{*}{sample 16} & home & $-$ & $-$ & $-$ & $-$ & $-$ & $-$ & $0.00$ \\
 & & gfs\_b & $57.63 \pm 0.24$ & $27.77 \pm 0.64$ & $8.21 \pm 0.62$ & $132.20 \pm 3.00$ & $0.00 \pm 0.00$ & $9.09$ & $0.10$ \\
 & & dec & $57.31 \pm 0.03$ & $39.00 \pm 0.29$ & $5.72 \pm 0.10$ & $142.47 \pm 0.66$ & $0.00 \pm 0.00$ & $7.68$ & $0.12$ \\
 & & domus & $-$ & $-$ & $-$ & $-$ & $-$ & $-$ & $0.00$ \\
 & & citius & $55.31 \pm 0.82$ & $29.69 \pm 0.58$ & $6.28 \pm 0.24$ & $221.00 \pm 3.99$ & $0.00 \pm 0.00$ & $6.81$ & $0.13$ \\
 & & raid\_a & $56.89 \pm 0.09$ & $31.38 \pm 0.20$ & $8.37 \pm 0.35$ & $201.27 \pm 0.59$ & $0.00 \pm 0.00$ & $8.06$ & $0.11$ \\
 & & wsc8a & $56.27 \pm 0.03$ & $32.09 \pm 0.21$ & $9.37 \pm 0.31$ & $203.50 \pm 1.84$ & $0.00 \pm 0.00$ & $7.44$ & $0.12$ \\
 & & home\_b & $60.62 \pm 0.70$ & $29.24 \pm 0.06$ & $8.27 \pm 0.09$ & $275.70 \pm 6.91$ & $0.33 \pm 0.47$ & $11.64$ & $0.06$ \\
 & & raid\_b & $57.54 \pm 1.24$ & $36.64 \pm 0.70$ & $6.25 \pm 0.44$ & $273.67 \pm 9.39$ & $12.33 \pm 1.25$ & $8.13$ & $0.05$ \\
 & & rooms & $57.95 \pm 0.49$ & $34.88 \pm 0.26$ & $6.37 \pm 0.26$ & $233.53 \pm 0.12$ & $0.00 \pm 0.00$ & $8.67$ & $0.10$ \\
 & & flower & $56.64 \pm 0.14$ & $40.23 \pm 0.13$ & $5.16 \pm 0.06$ & $244.13 \pm 1.13$ & $0.00 \pm 0.00$ & $6.96$ & $0.13$ \\
 & & office & $51.34 \pm 0.13$ & $26.54 \pm 0.11$ & $7.65 \pm 0.16$ & $522.00 \pm 3.41$ & $0.00 \pm 0.00$ & $3.56$ & $0.22$ \\
 & & autolab & $53.26 \pm 0.17$ & $31.50 \pm 0.38$ & $6.19 \pm 0.04$ & $462.23 \pm 3.94$ & $0.00 \pm 0.00$ & $4.79$ & $0.17$ \\
 & & maze & $-$ & $-$ & $-$ & $-$ & $-$ & $-$ & $0.00$ \\
 & & hospital & $52.54 \pm 0.13$ & $28.57 \pm 0.15$ & $6.71 \pm 0.19$ & $3359.17 \pm 13.58$ & $0.00 \pm 0.00$ & $4.43$ & $0.18$ \\
\hline
\end{tabular}
\end{table*}

\begin{table*}
\centering
\small
\caption{\label{Tab:average_results} Average results ($x \pm \sigma$) for all simulated environments}
\begin{tabular}{|c|c|c|c|c|c|Hc|}

\hline
Alg. & Prepr. & Dist.(cm) & Vel.(cm/s) & Vel.ch.(cm/s) & \# Blockades & $error$ & $quality$\\
\hline
\hline

IQFRL & $-$ & $55.36 \pm 2.57$ & $28.34 \pm 3.60$ & $6.10 \pm 1.04$ & $0.00 \pm 0.00$ & $6.99 \pm 2.40$ & $0.14 \pm 0.05$ \\
\hline
\multirow{2}{*}{MOGUL} & min 16 & $54.42 \pm 1.99$ & $30.30 \pm 4.13$ & $5.54 \pm 1.05$ & $40.33 \pm 73.78$ & $5.94 \pm 1.86$ & $0.05 \pm 0.02$ \\
\cline{2-8}
 & sample 16 & $55.10 \pm 2.83$ & $31.02 \pm 5.76$ & $6.07 \pm 1.39$ & $6.42 \pm 15.78$ & $6.49 \pm 2.39$ & $0.08 \pm 0.03$ \\
\hline
\multirow{2}{*}{MPNN} & min 16 & $56.86 \pm 1.87$ & $29.59 \pm 5.09$ & $6.05 \pm 1.76$ & $39.39 \pm 55.37$ & $8.21 \pm 2.07$ & $0.03 \pm 0.03$ \\
\cline{2-8}
 & sample 16 & $67.30 \pm 7.88$ & $23.69 \pm 4.99$ & $6.64 \pm 0.76$ & $17.79 \pm 18.12$ & $18.20 \pm 7.39$ & $0.02 \pm 0.03$ \\
\hline
\multirow{2}{*}{$\nu$-SVR} & min 16 & $57.17 \pm 3.05$ & $29.79 \pm 4.83$ & $7.07 \pm 2.05$ & $0.85 \pm 1.88$ & $8.47 \pm 2.49$ & $0.07 \pm 0.06$ \\
\cline{2-8}
 & sample 16 & $56.11 \pm 2.49$ & $32.29 \pm 4.27$ & $7.05 \pm 1.23$ & $1.05 \pm 3.40$ & $7.27 \pm 2.13$ & $0.10 \pm 0.06$ \\
\hline

\end{tabular}
\end{table*}

\begin{table*}[tb!]
\small
\centering
\caption{\label{Tab:real_results}Average results ($x \pm \sigma$) of IQFRL for the real environments}
\begin{tabular}{|c|c|c|c|c|c|c|c|c|}
\hline
Env. & Dist.(cm) & Vel.(cm/s) & Vel.ch.(cm/s) & Time(s) & $\#$ Blockades & $quality$ \\
\hline
\hline
real env 1 & $54.13 \pm 2.59$ & $19.86 \pm 1.52$ & $1.36 \pm 0.21$ & $100.70$ & $0.00 \pm 0.00$ & $0.13$\\
real env 2 & $59.29 \pm 2.74$ & $21.94 \pm 1.43$ & $1.72 \pm 2.50$ & $118.50$ & $0.00 \pm 0.00$ & $0.08$\\
\hline
\end{tabular}
\end{table*}

In general, all the algorithms except MPNN with \textit{Sample 16} preprocessing, produced a distance that is very close to the reference (between 40 cm and 60 cm to the wall at its right). Note that in cases where the best distance is very different from that obtained by IQFRL, this is because several blockades happened. Therefore, those controllers have the advantage of being continually repositioned into the perfect situation. The best results in speed are those obtained by $\nu$-SVR and MOGUL but, in general, due to a worsening in the distance to the wall or an increase in the number of blockades. The same applies to the speed change. In those cases where it is too low, like in some cases for MOGUL or MPNN, the robot is not able to trace some curves safely. IQFRL is the approach that gets the best quality values, reflecting not only the adequate values for the distance, velocity and smoothness in all the environments but, also, its robustness: it is the unique approach that never blocked or failed to complete the laps in any of the environments. 

In order to compare the experimental results, non-parametric tests of multiple comparisons have been used. Their use is recommended in those cases in which the objective is to compare the results of a new algorithm against various methods simultaneously. The Friedman test with Holm post-hoc test was selected as the method for detecting significant differences among the results. The test is performed for the $quality$ indicator in table \ref{Tab:results}.

The statistical test (table \ref{Tab:quality}) shows that the difference of the quality of the IQFRL approach is statistically significant. Only $\nu$-SVR and MOGUL with sample 16 preprocessing are comparable to IQFRL, as the number of blockades is very low or null in some environments. 

Additionally, table \ref{Tab:real_results} shows the results obtained by IQFRL in two real environments. As in the previous tables, the results are the average and standard deviation over 5 laps. The distance to the wall is lower than 60 cm, showing a good behavior,  although the velocity seems to be low, this is because corners are very close to each other and the robot does not have time to accelerate. Also, the velocity change reflects a very smooth movement as changes in velocity take more time in the real robot.

\begin{table}[b!]
\centering
\small
\caption{\label{Tab:eswa_quality}Non-parametric test for $quality$ of table \ref{Tab:eswa}.}
\begin{tabular}{|c|c|c|}
\hline
Alg. & Ranking & Holm \textit{p}-value\\
\hline
\hline
IQFRL & $2.9$ & $-$\\
\hline
COR & $3.9$ & $0.01$\\
\hline
WCOR & $3.7$ & $0.017$\\
\hline
HSWLR & $2.8$ & $0.05$\\
\hline
TSK & $1.7$ & $0.006$\\
\hline
\multicolumn{3}{|c|}{Friedman \textit{p}-value = $0.19$}\\
\multicolumn{3}{|c|}{Holm's rejects hypothesis with \textit{p}-value $<= 0.005$}\\
\hline
\end{tabular}
\end{table}

\begin{table*}[tb!]
\centering
\small
\caption{\label{Tab:eswa}Average results ($x \pm \sigma$) of IQFRL and several approaches with preprocessing based on expert knowledge \cite{Mucientes10_eswa}}
\begin{tabular}{|c|c|c|c|c|c|c|Hc|}
\hline
Alg. & Env. & Dist.(cm) & Vel.(cm/s) & Vel.ch.(cm/s) & Time(s) & \# Blockades & $error$ & $quality$\\
\hline
\hline
\multirow{5}{*}{IQFRL}
 & wsc8a & $56.96 \pm 1.00$ & $27.45 \pm 0.84$ & $7.70 \pm 0.33$ & $233.10 \pm 5.28$ & $0.00 \pm 0.00$ & $8.52$ & $0.11$ \\
 & rooms & $57.38 \pm 0.34$ & $30.97 \pm 0.34$ & $6.38 \pm 0.43$ & $261.93 \pm 4.60$ & $0.00 \pm 0.00$ & $8.54$ & $0.10$ \\
 & autolab & $52.91 \pm 0.20$ & $28.75 \pm 0.31$ & $5.57 \pm 0.48$ & $499.33 \pm 9.74$ & $0.00 \pm 0.00$ & $4.74$ & $0.17$ \\
 & office & $51.37 \pm 0.57$ & $24.20 \pm 0.18$ & $6.65 \pm 0.25$ & $578.27 \pm 2.92$ & $0.00 \pm 0.00$ & $3.81$ & $0.21$ \\
 & hospital & $51.09 \pm 0.19$ & $26.68 \pm 0.10$ & $6.18 \pm 0.35$ & $3608.07 \pm 21.72$ & $0.00 \pm 0.00$ & $3.31$ & $0.23$ \\
\hline
\multirow{5}{*}{COR}
 & wsc8a & $53.20 \pm 1.33$ & $39.86 \pm 0.71$ & $5.67 \pm 0.83$ & $174.98 \pm 1.79$ & $0.00 \pm 0.00$ & $4.89$ & $0.17$\\
 & rooms & $46.80 \pm 0.59$ & $37.82 \pm 0.41$ & $6.76 \pm 0.31$ & $227.16 \pm 1.03$ & $0.00 \pm 0.00$ & $5.10$ & $0.16$\\
 & autolab & $56.88 \pm 0.91$ & $25.69 \pm 0.79$ & $10.79 \pm 0.21$ & $587.96 \pm 39.72$ & $0.00 \pm 0.00$ & $9.62$ & $0.09$\\
 & office & $55.97 \pm 1.65$ & $32.48 \pm 0.90$ & $4.06 \pm 0.28$ & $457.58 \pm 15.00$ & $0.00 \pm 0.00$ & $8.13$ & $0.11$\\
 & hospital & $54.12 \pm 0.92$ & $35.63 \pm 0.77$ & $6.95 \pm 0.28$ & $2864.92 \pm 45.27$ & $0.00 \pm 0.00$ & $6.15$ & $0.14$\\
\hline
\multirow{5}{*}{WCOR}
 & wsc8a & $52.79 \pm 1.36$ & $36.98 \pm 1.85$ & $7.37 \pm 0.62$ & $187.90 \pm 9.78$ & $0.00 \pm 0.00$ & $4.81$ & $0.17$\\
 & rooms & $51.17 \pm 0.77$ & $37.19 \pm 0.27$ & $9.15 \pm 0.24$ & $234.04 \pm 2.70$ & $0.00 \pm 0.00$ & $3.33$ & $0.23$\\
 & autolab & $52.97 \pm 1.10$ & $33.47 \pm 0.89$ & $7.12 \pm 0.52$ & $455.98 \pm 41.60$ & $0.00 \pm 0.00$ & $5.33$ & $0.16$\\
 & office & $54.59 \pm 1.10$ & $33.13 \pm 0.97$ & $6.76 \pm 0.53$ & $448.16 \pm 10.36$ & $0.00 \pm 0.00$ & $6.82$ & $0.13$\\
 & hospital & $55.26 \pm 1.01$ & $33.71 \pm 0.14$ & $6.52 \pm 0.12$ & $3073.98 \pm 23.63$ & $0.00 \pm 0.00$ & $7.36$ & $0.12$\\
\hline
\multirow{5}{*}{HSWLR}
 & wsc8a & $51.42 \pm 0.78$ & $30.46 \pm 1.01$ & $3.36 \pm 0.13$ & $222.34 \pm 6.09$ & $0.00 \pm 0.00$ & $4.23$ & $0.19$\\
 & rooms & $50.09 \pm 0.88$ & $28.71 \pm 0.29$ & $3.04 \pm 0.20$ & $290.70 \pm 3.66$ & $0.00 \pm 0.00$ & $3.21$ & $0.24$\\
 & autolab & $51.50 \pm 0.34$ & $23.50 \pm 0.97$ & $3.05 \pm 0.14$ & $618.40 \pm 20.98$ & $0.00 \pm 0.00$ & $5.00$ & $0.17$\\
 & office & $53.43 \pm 1.22$ & $24.69 \pm 0.66$ & $3.73 \pm 0.11$ & $594.74 \pm 13.16$ & $0.00 \pm 0.00$ & $6.62$ & $0.13$\\
 & hospital & $54.60 \pm 1.65$ & $25.07 \pm 0.49$ & $3.89 \pm 0.06$ & $4209.68 \pm 166.14$ & $0.00 \pm 0.00$ & $7.63$ & $0.12$\\
\hline
\multirow{5}{*}{TSK}
 & wsc8a & $51.43 \pm 1.36$ & $37.54 \pm 1.53$ & $5.20 \pm 0.50$ & $182.54 \pm 8.35$ & $0.00 \pm 0.00$ & $3.53$ & $0.22$\\
 & rooms & $49.07 \pm 1.08$ & $37.05 \pm 0.82$ & $4.96 \pm 0.21$ & $227.58 \pm 4.46$ & $0.00 \pm 0.00$ & $3.13$ & $0.24$\\
 & autolab & $51.87 \pm 2.99$ & $33.05 \pm 1.33$ & $4.61 \pm 0.11$ & $465.56 \pm 15.33$ & $0.00 \pm 0.00$ & $4.38$ & $0.19$\\
 & office & $53.75 \pm 0.97$ & $34.26 \pm 0.65$ & $5.24 \pm 0.22$ & $432.38 \pm 10.48$ & $0.00 \pm 0.00$ & $5.95$ & $0.14$\\
 & hospital & $54.50 \pm 1.49$ & $34.31 \pm 0.32$ & $5.01 \pm 0.11$ & $3053.74 \pm 123.72$ & $0.00 \pm 0.00$ & $6.62$ & $0.13$\\
\hline
\end{tabular}
\end{table*}

Finally, the IQFRL proposal was compared with the proposals presented in \cite{Mucientes10_eswa} for learning rules for the wall-following behavior. The purpose of this comparison is to check if IQFRL is competitive against other methods which use expert knowledge for sensor data preprocessing. Four different approaches were used: the COR methodology, the weighted COR methodology (WCOR), Hierarchical Systems of Weighted Linguistic Rules (HSWLR) and a local evolutionary learning of Takagi-Sugeno rules (TSK). For these approaches, four input variables were defined by an expert: right distance, left distance, velocity, and the orientation (alignment) of the robot to the wall at its right. Moreover, the granularities of each variable were also defined by the expert. Table \ref{Tab:eswa} presents the comparison between these approaches and the IQFRL proposal on those environments which are common.

The IQFRL approach exhibited the highest quality in the two most complex environments (office and hospital). Moreover, table \ref{Tab:eswa_quality} shows the non-parametric tests performed over $quality$. The Friedman p-value is higher than in table 9, due to the low number of environments available for comparisons. As can be seen, there is no statistically significant difference regarding the $quality$. That is, the controllers learned with embedded preprocessing has similar performance to the methods that use expert knowledge to preprocess the data. 

\subsection{Complexity of the Rules}
An example of a rule learned by IQFRL is presented in Fig. \ref{Fig:rule}. The antecedent part is composed of a single QFP. The linguistic value $A_{d}^{5,\ 1}$ indicates a low distance, while $A_{b}^{4,\ 1}$ denotes that the beams sector of the proposition is formed by the frontal and right parts of the robot. Therefore, the rule describes a situation where the robot is too close to the wall and, if it continues, it will collide. Because of that, the consequent indicates a zero linear velocity and a turn of the robot to the left, in order to get away from the wall without getting the robot into risk.

\begin{figure}[!h]
\centering
\fbox{
\parbox{0.42\textwidth}{
\small
IF\\
\hspace*{1cm}$d(h)$ is $A_{d}^{5,\ 1}$ in $50$ percent of $A_{b}^{4,\ 1}$\\
THEN\\
\hspace*{1cm}$vlin$ is $A_{vlin}^{1}$ and\\
\hspace*{1cm}$vang$ is $A_{vang}^{19}$
}
}
\caption{\label{Fig:rule}A typical rule learned by IQFRL. $A_{d}^{5,\ 1}$ indicates a low distance and $A_{b}^{4,\ 1}$ indicates the frontal and right sectors.}
\end{figure}

Table \ref{Tab:number} shows the number of rules learned for the different situations by each of the methods based on rules. MOGUL is implemented as a multiple-input single-output (MISO) algorithm, therefore for each output, different rule bases were learned. Moreover, table \ref{Tab:gran} shows the complexity of the learned rules in terms of mean and standard deviation of the number of propositions and granularities for each input variable.

\begin{table*}[tb!]
\centering
\small
\caption{\label{Tab:number}Number of rules learned}
\begin{tabular}{|c|c|c|c|c|c|}
\hline
Alg. & Preproc. & Output & \#$R_{straight}$ & \#$R_{convex}$ &\#$R_{concave}$\\
\hline
\hline
IQFRL & $-$ & Both & $108.00 \pm 18.88$ & $47.80 \pm 16.09$ & $40.40 \pm 10.65$\\
\hline
\multirow{4}{*}{MOGUL} & min 16 &  \textit{vlin}  &  $548.60 \pm 25.60$  &  $308.20 \pm 12.12$  &  $680.20 \pm 24.43$ \\
& & \textit{vang}  &  $547.00 \pm 16.37$  &  $302.80 \pm 21.57$  &  $712.40 \pm 23.79$ \\
\cline{2-6}
 & sample 16 &  \textit{vlin}  &  $507.80 \pm 29.88$  &  $268.20 \pm 12.66$  &  $664.80 \pm 8.52$ \\
& & \textit{vang}  &  $530.20 \pm 26.48$  &  $252.80 \pm 8.28$  &  $709.80 \pm 34.19$ \\
\hline
\end{tabular}
\end{table*}

\begin{table*}[tb!]
\centering
\small
\caption{\label{Tab:gran}Complexity of the rules}
\begin{tabular}{|c|c|c|c|c|c|c|c|}
\hline
Alg. & Preproc. & Dataset & Output & Propositions & $g_{d}$ &$g_{b}$ & $g_{v}$\\
\hline
\hline 
\multirow{3}{*}{IQFRL} & \multirow{3}{*}{$-$} & Straight & \multirow{3}{*}{Both} & $2.74 \pm 0.94$ & $7.02 \pm 10.52$ & $5.98 \pm 5.62$ & $6.21 \pm 1.53$\\
& & Convex &  & $2.68 \pm 0.69$ & $15.37 \pm 23.59$ & $11.22 \pm 8.50$ & $6.55 \pm 1.03$\\
& & Concave &  & $2.78 \pm 1.18$ & $3.80 \pm 1.79$ & $7.07 \pm 6.86$ & $6.16 \pm 1.42$\\
\hline
\multirow{12}{*}{MOGUL}
& \multirow{6}{*}{min 16} & \multirow{2}{*}{Straight} & \textit{vlin} & $17.00 \pm 0.00$ & $24.35 \pm 109.80$ & $16.00 \pm 0.00$ & $39.44 \pm 137.45$\\
&  &  & \textit{vang} & $17.00 \pm 0.00$ & $24.49 \pm 107.66$ & $16.00 \pm 0.00$ & $35.19 \pm 117.75$\\
\cline{3-8}
&  & \multirow{2}{*}{Convex} & \textit{vlin} & $17.00 \pm 0.00$ & $32.34 \pm 125.75$ & $16.00 \pm 0.00$ & $51.68 \pm 172.27$\\
&  &  & \textit{vang} & $17.00 \pm 0.00$ & $38.99 \pm 144.86$ & $16.00 \pm 0.00$ & $45.07 \pm 146.38$\\
\cline{3-8}
&  & \multirow{2}{*}{Concave} & \textit{vlin} & $17.00 \pm 0.00$ & $22.93 \pm 100.76$ & $16.00 \pm 0.00$ & $32.79 \pm 106.77$\\
&  &  & \textit{vang} & $17.00 \pm 0.00$ & $23.35 \pm 103.39$ & $16.00 \pm 0.00$ & $37.56 \pm 122.75$\\
\cline{2-8}
& \multirow{6}{*}{sample 16} & \multirow{2}{*}{Straight} & \textit{vlin} & $17.00 \pm 0.00$ & $26.23 \pm 117.41$ & $16.00 \pm 0.00$ & $33.98 \pm 108.16$\\
&  &  & \textit{vang} & $17.00 \pm 0.00$ & $26.25 \pm 116.18$ & $16.00 \pm 0.00$ & $37.60 \pm 126.86$\\
\cline{3-8}
&  & \multirow{2}{*}{Convex} & \textit{vlin} & $17.00 \pm 0.00$ & $25.68 \pm 103.29$ & $16.00 \pm 0.00$ & $49.61 \pm 160.18$\\
&  &  & \textit{vang} & $17.00 \pm 0.00$ & $31.06 \pm 119.50$ & $16.00 \pm 0.00$ & $46.56 \pm 151.27$\\
\cline{3-8}
&  & \multirow{2}{*}{Concave} & \textit{vlin} & $17.00 \pm 0.00$ & $23.50 \pm 105.79$ & $16.00 \pm 0.00$ & $33.62 \pm 112.09$\\
&  &  & \textit{vang} & $17.00 \pm 0.00$ & $23.95 \pm 106.27$ & $16.00 \pm 0.00$ & $34.63 \pm 121.52$\\
\hline
\end{tabular}
\end{table*}

The IQFRL approach is able to learn knowledge bases with a much lower number of rules than MOGUL, even though it is learning both outputs at the same time. The learning of QFRs results in a low number of propositions per rule, thus demonstrating its generalization ability, in spite of the huge input space dimensionality. Moreover, the granularities of each of the input variables are, in general, also low. Therefore, the learned knowledge bases show a low complexity without losing accuracy.

\section{Real World Applications\label{Sec:realWorld}}
Two of the most used behaviors in mobile robotics are path and object tracking. In recent years several real applications of these behaviors have been described in the literature in different realms. For instance, in \cite{kim2004autonomous}, a tour-guide robot that can either follow a predefined route or a tour-guide person was shown. With a similar goal, an intelligent hospital service robot was presented in \cite{shieh2004design}. In this case, the robot can improve the services provided in the hospital through autonomous navigation based on following a path. More recently, in \cite{lopez2013watchbot} a team of robots that cooperate in a building developing maintenance and surveillance tasks was presented.

More dynamic environments were described in \cite{gamallo2010omnivision,kummerle2013navigation}, where the robot had to operate in buildings and populated urban areas. These environments introduce numerous challenges to autonomous mobile robots as they are highly complex. Finally, in \cite{Gonzalez-Sieira13_robot} the authors presented a motion planner that was able to generate paths taking into account the uncertainty due to controls and measurements.

In these and other real applications, the robot has to deal with static and moving objects, including the presence of people surrounding the robot, etc. All these difficulties make necessary the combination of behaviors to perform tasks like path or people tracking in real environments. In order to implement these tasks in a safe way, the robot must be endowed with the ability to avoid collisions with all the objects in the environment while implementing the tasks. These behaviors are challenging tasks that allow us to show the performance of the IQFRL-based approach in realistic conditions. The following behaviors are considered in this section, in order of increasing complexity:
\begin{enumerate}
 \item \textit{Path tracking with obstacles avoidance}. In this behavior, the mobile robot must follow a path with obstacles in it. A typical application of this behavior is a tour-guide robot that has to follow a predefined tour in a museum. Although in the initial path there were no obstacles in the trajectory, the modification of the environment with new exhibitors and the presence of people make it necessary that the robot modify the predefined route, avoiding the collision with the obstacles and returning to the predefined path as quickly as possible.
 \item \textit{Object tracking with fixed obstacles avoidance}. In this case, the robot has to follow the path of a moving object while being at a reference distance to the object. For instance, a tour-guide person being followed by a robot with extended information on a screen. If the followed object comes too close to an obstacle, the robot must avoid the collision while maintaining the tracking behavior.
 \item \textit{Object tracking with moving obstacle avoidance}. This behavior is a modification of the previous one, and presents a more difficult problem. In addition to the fixed obstacles avoidance, the robot has to track an object while preventing collisions with moving obstacles that are crossing between the robot and the tracked object. These moving obstacles can be persons walking around or even other mobile robots doing their own behaviors.
\end{enumerate}

In order to perform these behaviors, a fusion of two different controllers has been developed. On one hand, a tracking controller \cite{Mucientes07_tfs} was used in order to follow the path or the moving object. On the other hand, the wall-following controller learned with the IQFRL algorithm was used as the collision avoidance behavior. Section \ref{sec:comparison} showed that this controller is robust and operates safely while performing the task. There were no blockades during the behavior in all the tests, neither from collisions nor from other reasons. The way in which the wall-following behavior is used in order to avoid collisions is: given an obstacle that is too close to the robot, it can be surrounded following the border of this obstacle in order to avoid a collision with it. The controller described in this paper follows the wall on its right, while for this task, the obstacle can be on both sides. This can easily be solved by a simple permutation of the laser beams depending on which side the 
obstacle is detected.

The wall-following behavior is only executed when the robot is too close to an object ---a value of 0.4 m has been used as threshold. The objective of the controller is to drive the robot to a state in which there is no danger of collision ---a value of 0.5 m has been established as a safe distance. As long as the robot is in a safe state the tracking behavior is resumed. This behavior controls the linear and angular velocities of the robot in order to place it at an objective point in every control cycle. This point is defined using the desired distance between the robot and the moving object. The tracking controller uses four different input variables:
\begin{itemize}
 \item The distance between the robot and the objective point:
 \begin{equation}
    d = \frac{\sqrt{(x_r-x_{obj})^2 + (y_r - y_{obj})^2}}{d_{ref}}
 \end{equation}
 where $(x_r, y_r)$ are the coordinates of the robot, $(x_{obj}, y_{obj})$ are the coordinates of the objective point and $d_{ref}$ is the reference distance between the robot and the objective point.
 \item The deviation of the robot with respect to the objective point:
 \begin{equation}
    dev = \operatorname{arctan}\left(\frac{y_{obj} - y_r}{x_{obj} - x_r}\right) - \theta_r
 \end{equation}
 where $\theta_r$ is the angle of the robot. A negative value of the deviation indicates that the robot is moving in a direction to the left of the objective point, while a positive value means that it is moving to the right.
 \item The difference of velocity between the robot and the objective point:
 \begin{equation}
  \Delta v = \frac{v_r - v_m}{v_{max}}
 \end{equation}
  where $v_r$, $v_m$ and $v_{max}$ are the linear velocities of the robot, the moving object, and the maximum velocity attainable by the robot.
  \item The difference in angle between the object and the robot:
  \begin{equation}
   \Delta \theta = \theta_m - \theta_r
  \end{equation}
  where $\theta_m$ is the angle of the moving object.
\end{itemize}

The reference distance ($d_{ref}$) is different depending on the type of behavior. For the path tracking behavior, there is no moving object tracking and, therefore, the robot follows the path with $d_{ref} = 0$ in order to do a perfect path tracking. In the other two behaviors the robot follows a moving object, so it is necessary to keep a safe distance ---a value of $d_{ref} = 0.5$ m was used in the experiments shown in this section.

\begin{figure*}[htb!]
\centering
\begin{tabular}{>{\centering\arraybackslash}m{0.35\textwidth}>{\centering\arraybackslash}m{0.65\textwidth}}
\subfigure[\emph{Path tracking with obstacles avoidance in \textit{M1}.}\label{Fig:pt_museum}]{\includegraphics[width=0.65\columnwidth]{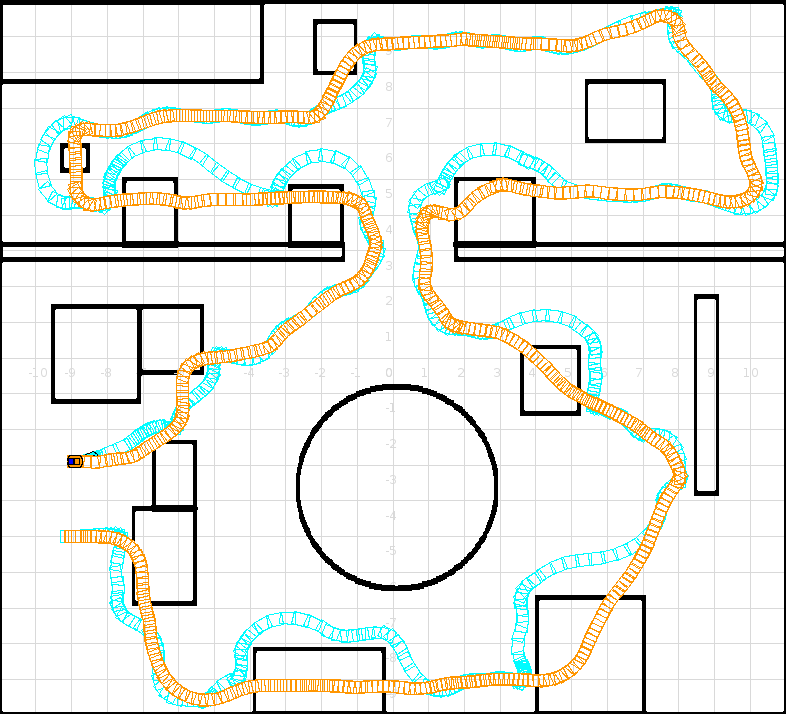}} &
\subfigure[\emph{Path tracking with obstacles avoidance in \textit{Domus}.}\label{Fig:pt_domus}]{\includegraphics[width=0.98\columnwidth]{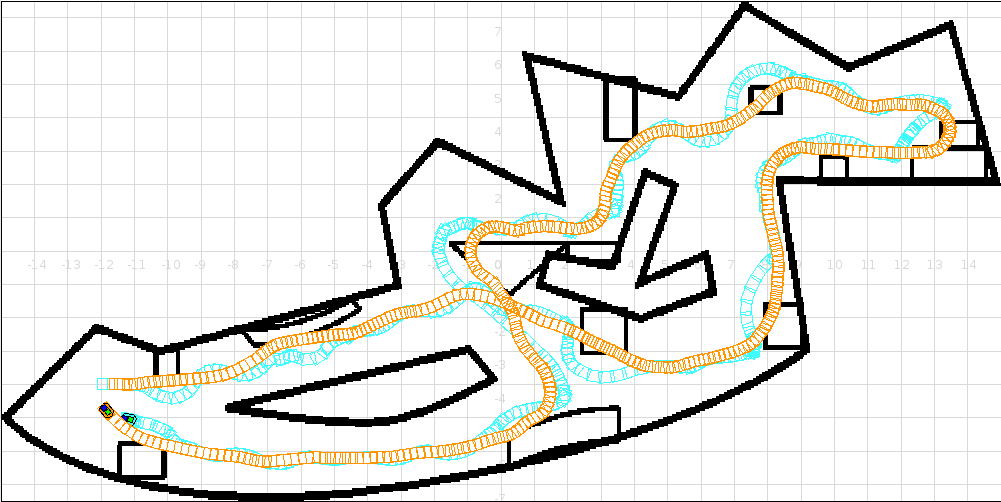}}\\
\subfigure[\emph{Object tracking with fixed obstacles avoidance in \textit{M1}.}\label{Fig:off_museum}]{\includegraphics[width=0.65\columnwidth]{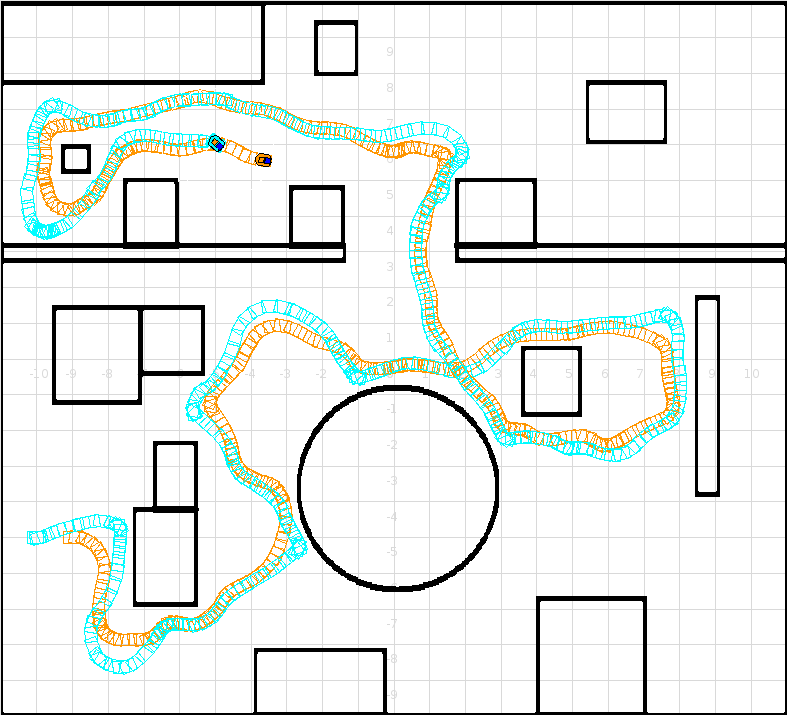}} &
\subfigure[\emph{Object tracking with fixed obstacles avoidance in \textit{Domus}.}\label{Fig:off_domus}]{\includegraphics[width=0.98\columnwidth]{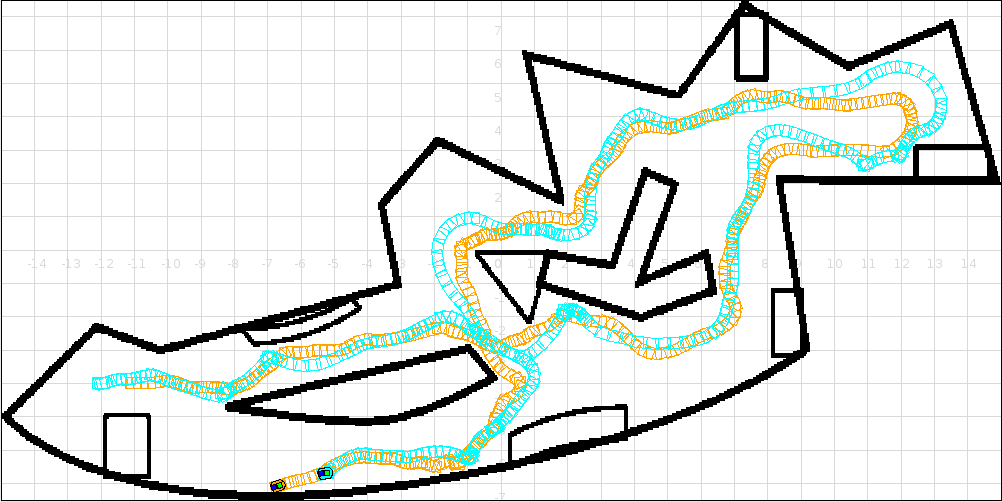}}\\
\subfigure[\emph{Object tracking with moving obstacle avoidance in \textit{M1}.}\label{Fig:ofm_museum}]{\includegraphics[width=0.65\columnwidth]{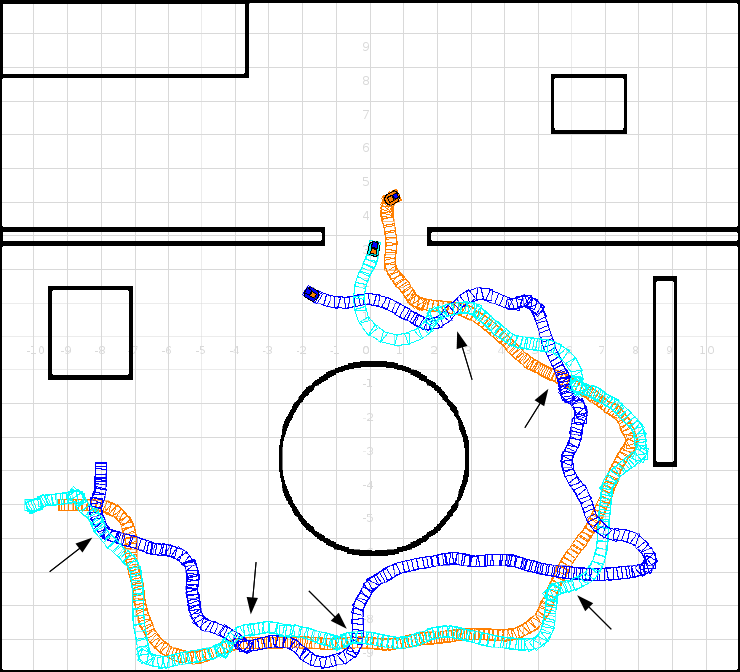}} &
\subfigure[\emph{Object tracking with moving obstacle avoidance in \textit{Domus}.}\label{Fig:ofm_domus}]{\includegraphics[width=0.98\columnwidth]{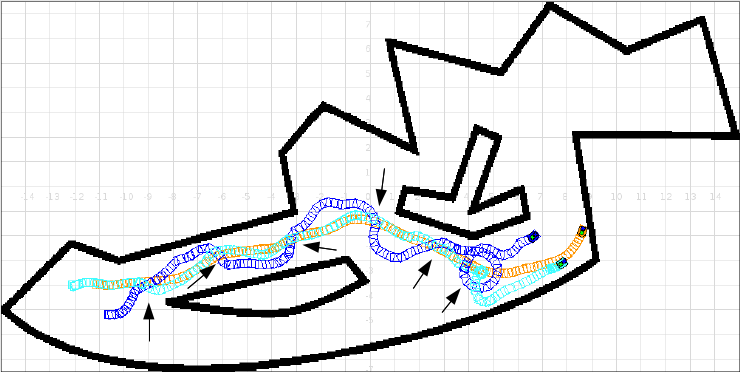}}\\
\end{tabular}
\caption{\label{Fig:real_applications}Experiments on real applications. Colors code: 1) Original path to be tracked in orange (medium grey); 2) Robot path in cyan (light grey); 3) Moving obstacle path in blue (dark grey). The arrows along the path in Figs. 20(e) and 20(f) indicate the places in which the moving obstacle interferes with the robot.}
\end{figure*}

The three behaviors have been validated in two different environments (\emph{M1} and \emph{Domus}) which try to reproduce the plant of a museum (Fig. \ref{Fig:real_applications}). Figs. \ref{Fig:pt_museum} and \ref{Fig:pt_domus} show the path tracking with obstacles avoidance behavior. The orange (medium grey) path represents the trajectory that has to be followed by the robot. This path also includes information of the velocity that the robot should have at each point. The higher the concentration of marks, the lower the linear velocity in that point of the path. Moreover, the path was generated without obstacles and, once the obstacles were added to the environment, the robot was placed at the beginning of the path in order to track it. The cyan (light grey) path indicates the trajectory implemented by the robot using the proposed combination of controllers (wall-following and tracking). It can be seen that the robot avoids successfully all the obstacles in its path, i.e., the wall following behavior deviates the robot from the predefined path when an obstacle generates a possibility of collision. When the robot overcomes the obstacle, it returns to the predefined path as quickly as possible.

In the case of the moving object tracking with fixed obstacles avoidance behavior (Figs. \ref{Fig:off_museum} and \ref{Fig:off_domus}), the cyan (light grey) line represents the path of the robot due to the combination of the controllers. Also, the orange (medium grey) path shows the trajectory of the moving object tracked by the robot. In this behavior, the moving object goes too close to some obstacles in several situations, forcing the controller to execute the wall following behavior in order to avoid collisions. Moreover, the wall-following controller is also executed when the moving object turns the corners very close to the obstacles, at a distance that is unsafe for the robot.

The last and most complex behavior is moving object tracking with moving obstacle avoidance (Figs. \ref{Fig:ofm_museum} and \ref{Fig:ofm_domus}). The cyan (light grey) path shows, once again, the path followed by the robot when it tracks the moving object (orange / medium grey path) while avoiding static and moving obstacles. Also, the path followed by the moving obstacle that should be avoided by the robot is shown in blue (dark grey). The arrows along the path indicate the places in which the obstacle interferes with the robot. This behavior shows the ability of the controller learned with the IQFRL algorithm to avoid collisions, even when the moving obstacle tries to force the robot to fail: the controller can detect the situation and perform the task safely, avoiding collisions.

\section{Conclusions\label{Sec:conclusions}}
This paper describes a new algorithm which is able to learn controllers with embedded preprocessing for mobile robotics. The transformation of the low-level variables into high-level variables is done through the use of Quantified Fuzzy Propositions and Rules. Furthermore, the algorithm involves linguistic labels defined by multiple granularity without limiting the granularity levels. The algorithm was extensively tested with the wall-following behavior both in several simulated environments and on a \textit{Pioneer 3-AT} robot in two real environments. The results were compared with some of the most well-known algorithms for learning controllers in mobile robotics. Non-parametric significance tests have been performed, showing a very good and a statistically significant performance of the IQFRL approach.

\section{Acknowledgements\label{Sec:acknowledgements}}
This work was supported by the Spanish Ministry of Economy and Competitiveness under grants TIN2011-22935 and TIN2011-29827-C02-02. I. Rodriguez-Fdez is supported by the Spanish Ministry of Education, under the FPU national plan (AP2010-0627). M. Mucientes is supported by the Ram\'{o}n y Cajal program of the Spanish Ministry of Economy and Competitiveness. This work was supported in part by the European Regional Development Fund (ERDF/FEDER) under the projects CN2012/151 and CN2011/058 of the Galician Ministry of Education.

NOTICE: this is the author’s version of a work that was accepted for publication in Applied Soft Computing. Changes resulting from the publishing process, such as peer review, editing, corrections, structural formatting, and other quality control mechanisms may not be reflected in this document. Changes may have been made to this work since it was submitted for publication. A definitive version was subsequently published in Applied Soft Computing, 26:123-142, 2015, doi:10.1016/j.asoc.2014.09.021.

\appendix

\section{IQFRL for Classification (IQFRL-C)\label{Sec:class}}
This section describes the modifications that are necessary to accomplish for adapting the IQFRL algorithm for classification problems.

\subsection{Examples and Grammar}
The structure of the examples used for classification is very similar to the one described in expression \ref{Eq:example}:

\begin{equation}\label{Eq:example_class}
e^l = \left( d\left( 1 \right), \: \ldots,  \: d\left( N_b\right), \: \textnormalit{velocity}, \: \textnormalit{class} \right)
\end{equation}
where $class$ represents the class of the example.

Furthermore, the consequent production (production 3) of the grammar (Fig. \ref{Fig:grammar}) must be modified to:
\begin{itemize}
\item[3.] consequent $\longrightarrow$ $F_c$
\end{itemize}
where $F_c$ is the linguistic label of the class. The output variable ($class$) has a granularity $g^{\# class}_c$.

\subsection{Initialization}
The consequent of the rules is initialized as $F_c = A^{\gamma}_c$ where $\gamma$ is the class that represents the example. Only those examples whose class is different from the default class ($A^{f}_c$) are used in the initialization of a new individual.
 
\subsection{Evaluation}
For each individual (rule) of the population, the following values are calculated:

\begin{itemize}

\item True positives ($\textnormalit{tp}$):
\begin{itemize}

\item $\#\textnormalit{tp} = \left|\left\{ e^l  \: :  \: C_l = C_j \wedge \textnormalit{DOF}_{j} \left( e^l \right) > 0 \right\}\right|$, where $C_l$ is the class of example $e^l$, $C_j$ is the class in the consequent of the $j$-th rule, and $\textnormalit{DOF}_{j} \left( e^l \right)$ is the $\textnormalit{DOF}$ of the $j$-th rule for the example $e^l$. $\#\textnormalit{tp}$ represents the number of examples that have been correctly classified by the rule.

\item $\textnormalit{tpd} = \sum_l \textnormalit{DOF}_{j} \left( e^l \right)  \: :  \: C_l = C_j$, i.e., the sum of the $\textnormalit{DOF}$s of the examples contributing to $\#\textnormalit{tp}$.

\item $\textnormalit{tp} = \#\textnormalit{tp} + \textnormalit{tpd}/\#\textnormalit{tp}$

\end{itemize}

\item False positives ($\textnormalit{fp}$):
\begin{itemize}

\item $\#\textnormalit{fp} = \left|\left\{ e^l  \: :  \: C_l \neq C_j \wedge \textnormalit{DOF}_{j} \left( e^l \right) > 0 \right\}\right|$: number of patterns that have been classified by the rule but belong to a different class.

\item $\textnormalit{fpd} = \sum_l \textnormalit{DOF}_{j} \left( e^l \right)  \: :  \: C_l \neq C_j$, i.e., the sum of the $\textnormalit{DOF}$s of the patterns that contribute to $\#\textnormalit{fp}$.

\item $\textnormalit{fp} = \#\textnormalit{fp} + \textnormalit{fpd}/\#\textnormalit{fp}$

\end{itemize}

\item False negatives ($\textnormalit{fn}$):
\begin{itemize}

\item $\#\textnormalit{fn} = n_{\textnormalit{ex}}^{C_j} - \#\textnormalit{tp}$, where $n_{\textnormalit{ex}}^{C_j} = \left|\left\{ e^l  \: :  \: C_l = C_j \right\}\right|$. $\#\textnormalit{fn}$ is the number of examples that have not been classified by the rule but belong to the class in the consequent of the rule.

\end{itemize}
\end{itemize}

The values of $\textnormalit{tp}$ and $\textnormalit{fp}$ take into account not only the number of examples that are correctly/incorrectly classified, but also the degree of fulfillment of the rule for each of the examples. In case that $\textnormalit{tpd} \approx 0$, then $\textnormalit{tp} \approx \#\textnormalit{tp}$, while if it is high ($\textnormalit{tpd} \approx \#\textnormalit{tp}$) then $\textnormalit{tp} \approx \#\textnormalit{tp} + 1$. Taking into account these definitions, the accuracy of an individual of the population can be described as:
\begin{equation}
\label{Eq:confidence-class}
\textnormalit{confidence} = \frac{1}{10^{\textnormalit{fp}}}
\end{equation}
 while the ability of generalization of a rule is calculated as:
\begin{equation}
\label{Eq:support}
\textnormalit{support} = \frac{\textnormalit{tp}}{\textnormalit{tp} + \#\textnormalit{fn}}
\end{equation} 

Finally, $\textnormalit{fitness}$ is defined as the combination of both values:
\begin{equation}
\label{Eq:rawFitness}
\textnormalit{fitness} = \textnormalit{confidence} \cdot \textnormalit{support}
\end{equation}
 which represents the strength of an individual.

\subsection{Mutation}
For classification, the probability that an example matches the output associated to a rule (Eq. \ref{Eq:probEx}) is binary. Therefore, in order to select the example ($e^{sel}$) that is going to be used for mutation, the following criteria is used:
\begin{itemize}
 \item For generalization, the probability for an example $e^l$ to be selected is:
\begin{equation}
 P(e^l = e^{sel}) = 1\ -\ \frac{\sum_j \textnormalit{DOF}_{j} \left( e^l \right) \cdot \textnormalit{confidence}_j}{\sum_j \textnormalit{DOF}_{j} \left( e^l \right)}
\end{equation}
where $\textnormalit{confidence}_j$ is the $\textnormalit{confidence}$ (Eq. \ref{Eq:confidence-class}) of the $j$-th individual. This probability measures the accuracy with which the individuals of the population cover the example $e^l$.
\item For specialization, the mutated individual uncovers the example $e^{sel}$. The probability to select $e^l$ for specialization is calculated as follow:
\begin{equation}
 P(e^l = e^{sel}) = 1\ -\ \textnormalit{DOF}_{j} \left( e^l \right)
\end{equation}

\end{itemize}

Finally, the consequent is mutated considering the class of the examples covered by the individual. Thus, the probability that the consequent of the individual $j$ change to the class $C_\gamma$ is defined as:

\begin{equation}
 P\left(j\ |\ C_\gamma\right) = \frac{\sum_l \textnormalit{DOF}_{j} \left( e^l \right)  \: :  \: C_l = C_\gamma}{\sum_l \textnormalit{DOF}_{j} \left( e^l \right)}
\end{equation}

\subsection{Performance}
The parameters used for IQFRL-C are the same as for regression (Sec. \ref{Sec:parameters}). Moreover, the default class is straight wall. Tables \ref{Tab:number_class} and \ref{Tab:gran_class} show the number of rules learned by the classification method IQFRL-C and the complexity of the rules learned in terms of mean and standard deviation of the number of propositions and granularities for each input variable. The number of rules for each situation is very low, resulting in very interpretable knowledge bases. Furthermore, the complexity of the rules is also low, as the number of propositions and granularities learned show that the rules are very general.

\begin{table}[tb!]
\centering
\small
\caption{\label{Tab:number_class}Number of rules learned for dataset by IQFRL-C}
\begin{tabular}{|c|c|c|c|c|}
\hline
\#$R_{straight}$ & \#$R_{convex}$ &\#$R_{concave}$\\
\hline
\hline
$-$ & $21.20 \pm 4.35$ & $10.00 \pm 1.41$\\
\hline
\end{tabular}
\end{table}

\begin{table}[tb!]
\centering
\small
\caption{\label{Tab:gran_class}Complexity of the rules learned by IQFRL-C}
\begin{tabular}{|c|c|c|c|}
\hline
Propositions & $g_{d}$ &$g_{b}$ & $g_{v}$\\
\hline
\hline
$2.67 \pm 0.90$ & $7.58 \pm 12.65$ & $8.41 \pm 8.49$ & $5.83 \pm 1.65$\\
\hline
\end{tabular}
\end{table}

Table \ref{Tab:confusion} shows the confusion matrix for the learned classifier. The matrix was obtained as the average of a 5-fold cross-validation over the sets. Moreover, the performance of the classifier was analyzed with the accuracy and the Cohen's $\kappa$ \cite{ben2007_eaia}. Both measures are very close to 1, showing the high performance of the classifier obtained with IQFRL-C.

\begin{table}[tb!]
\centering
\small
\caption{\label{Tab:confusion}Confusion matrix for the classifier}
\begin{tabular}{|c|c|c|c|}
\hline
Actual/Predicted & Straight & Convex & Concave\\
\hline
Straight & 30.85 & 2.40 & 0.23\\
\hline
Convex & 0.70 & 30.97 & 0.00\\
\hline
Concave & 0.23 & 0.06 & 34.55\\
\hline
\hline
\multicolumn{4}{|c|}{$\textnormalit{Accuracy} = 0.96$}\\
\multicolumn{4}{|c|}{Cohen's $\kappa = 0.94$}\\
\hline
\end{tabular}
\end{table}

\bibliographystyle{elsarticle-num}
\bibliography{biblio}

\end{document}